%% file: ms.tex
\documentclass[10pt,twocolumn,letterpaper]{article}
\PassOptionsToPackage{numbers,sort&compress}{natbib}
\usepackage[pagenumbers]{cvpr}

\usepackage[title]{appendix}
\usepackage{multirow}
\usepackage{amsmath}

\input{preamble}

\usepackage{tikz}
\usetikzlibrary{shapes.geometric}

\usepackage{graphicx}  
\usepackage{caption}
\usepackage{subcaption}
\usepackage{array} 
\definecolor{cvprblue}{rgb}{0.21,0.49,0.74}
\definecolor{nvgreen}{rgb}{0.463,0.725,0.}
\usepackage[pagebackref,breaklinks,colorlinks]{hyperref}
\usepackage{nicefrac}
\setlist[enumerate]{wide=0pt, widest=99,leftmargin=\parindent, labelsep=*}

\hypersetup{
    colorlinks,
    citecolor=nvgreen,
}
\usepackage{nimbusmononarrow}

\def\confName{CVPR}
\def\confYear{2025}

\input{macros.tex}

\title{Dynamic Camera Poses and Where to Find Them}

\author{Chris Rockwell$^{1,2}$ \qquad Joseph Tung$^3$  \qquad Tsung-Yi Lin$^1$ \\ Ming-Yu Liu$^1$ \qquad David F. Fouhey$^3$ \qquad Chen-Hsuan Lin$^1$ \vspace{0.2cm} \\
NVIDIA$^1$ \qquad University of Michigan$^2$ \qquad New York University$^3$
\vspace{2pt} \\
{\fontsize{10}{10}\selectfont \url{https://research.nvidia.com/labs/dir/dynpose-100k}}
}

\begin{document}

\twocolumn[{
    \renewcommand\twocolumn[1][]{#1}
    \maketitle
    \begin{center}
        \input{figures/teaser}
    \end{center}
}]

\input{sections/0_abstract}

\input{sections/1_intro}

\input{sections/2_related}
\input{sections/3_method}

\input{sections/4_datasets}

\input{sections/5_experiments}
\input{sections/6_conclusion}

\begingroup
\small

\input{ms.bbl}
\endgroup

\newcommand{\arxiv}{1}

\ifthenelse{\equal{\arxiv}{1}}
    {
        \clearpage
        \begin{appendices}
            The appendix contains additional experiments and detail.
            \name results are best viewed through video on the project webpage.
            Videos use 12 fps to correspond to camera poses, which are annotated at 12 fps.
            
            \input{supp/method_details}
            \input{supp/dataset_details}
            \input{supp/experimental_details}
            \input{supp/additional_dataset_results}

            \input{supp/additional_pose_results}
        \end{appendices}
    }
    {}
\end{document}

%% file: preamble.tex
\usepackage[dvipsnames]{xcolor}
\usepackage{amsmath}
\usepackage{amssymb}
\usepackage{amsthm}
\usepackage{soul}
\usepackage{bbm}
\usepackage[most]{tcolorbox}
\usepackage{enumitem}
\usepackage{ifthen}

\definecolor{dark_green}{rgb}{0, 0.5, 0}
\definecolor{dark_blue}{rgb}{0, 0, 0.5}

%% file: macros.tex
\newcommand{\parnobf}[1]{\vspace{4pt} \par \noindent {\bf {#1}.}}
\newcommand{\parnoit}[1]{\vspace{4pt} \par \noindent {\it {#1}.}}

\definecolor{ibm1}{HTML}{648FFF}
\definecolor{ibm2}{HTML}{DC267F}
\definecolor{ibm3}{HTML}{FE6100}
\definecolor{ibm4}{HTML}{FFB000}
\definecolor{ibm5}{HTML}{785EF0}
\definecolor{ibm6}{HTML}{88CCEE}

\newcommand{\name}{DynPose-100K\xspace}

\definecolor{recon_pts}{RGB}{46, 220, 255}
\definecolor{reproj_err}{RGB}{49, 224, 102}
\definecolor{vlm}{RGB}{31,119,180}

\definecolor{hands23}{RGB}{255,125,251}
\definecolor{flow}{RGB}{170, 28, 252}
\definecolor{tracking}{RGB}{65, 28, 252}
\definecolor{masking}{RGB}{50, 168, 86}
\definecolor{focal}{RGB}{207, 192, 31}
\definecolor{distort}{RGB}{247, 150, 13}
\definecolor{dynpose}{RGB}{255, 71, 71}

\definecolor{staryellow}{RGB}{255, 255, 0}

\newcommand{\yellowstar}{
  \tikz[baseline=-0.7ex] \node[star, star points=5, star point ratio=2.5, draw=black, fill=staryellow, line width=0.4pt, inner sep=1.0pt] {};
}
\renewcommand{\paragraph}[1]{\vspace{2pt} \noindent \textbf{#1}}

%% file: figures/teaser.tex
    \centering
    \includegraphics[width=\linewidth]{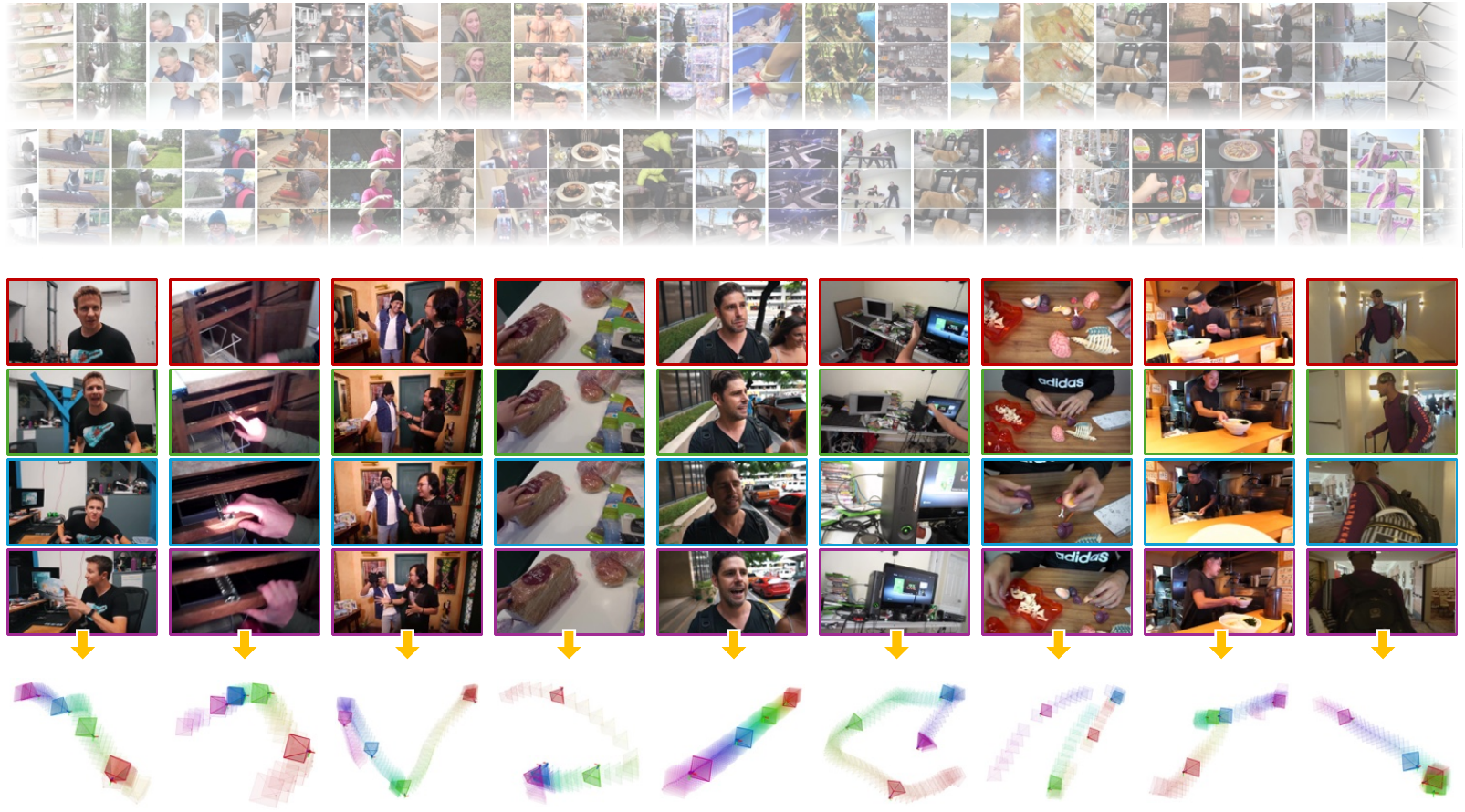}
    \captionof{figure}{
        We introduce \textbf{\name}, a large-scale video dataset of dynamic content with camera annotations. \name consists of 100,131 Internet videos that span diverse settings.
        We curate \name such that videos contain dynamic content while ensuring the cameras are able to be estimated (including intrinsics and poses).
        Towards this end, we address two challenging problems: (a) identifying the videos suitable for camera estimation, and (b) improving the camera estimation algorithm for dynamic videos.
    }
    \label{fig:teaser}
    \vspace{16pt}

%% file: sections/0_abstract.tex
\begin{abstract}
\vspace{-2mm}

Annotating camera poses on dynamic Internet videos at scale is critical for advancing fields like realistic video generation and simulation.
However, collecting such a dataset is difficult, as most Internet videos are unsuitable for pose estimation.
Furthermore, annotating dynamic Internet videos present significant challenges even for state-of-the-art methods.
In this paper, we introduce \name, a large-scale dataset of dynamic Internet videos annotated with camera poses.
Our collection pipeline addresses filtering using a carefully combined set of task-specific and generalist models.
For pose estimation, we combine the latest techniques of point tracking, dynamic masking, and structure-from-motion to achieve improvements over the state-of-the-art approaches.
Our analysis and experiments demonstrate that \name\ is both large-scale and diverse across several key attributes, opening up avenues for advancements in various downstream applications.

\end{abstract}
\vspace{-1.2em}

%% file: sections/1_intro.tex
\section{Introduction}
\label{sec:intro}

Annotating large-scale dynamic Internet video with camera information has the potential to advance many problems in computer vision and robotics.
Such a dataset could power generative models~\cite{brooks2024video,blattmann2023stable,ho2022imagen,singer2022make,ni2023conditional,geyer2023tokenflow,yang2023rerender,ni2023conditional} to create camera-controlled dynamic videos~\cite{wang2023motionctrl,he2024cameractrl,xu2024camco,kuang24,chen2024v3d,zhang2024tora,nvidia2025,ren2025gen3c} and 4D scenes~\cite{zhao2024genxd,bahmani2024tc4d,singer2023text}.
It could also enable large-scale training of view synthesis models~\cite{wang2021ibrnet,yu2021pixelnerf,deitke2024objaverse,wiles2020synsin,chan2022efficient,gao2022get3d,liu2023zero} for extended reality applications.
Additionally, this data could be transformational in robotics for tasks like imitation learning~\cite{bahl2022human,o2023open,fu2024mobile} or training in realistic simulation environments~\cite{blattmann2023align,hu2023gaia,wang2023drivedreamer}.

However, constructing a dataset of dynamic videos with camera annotations is challenging.
First, most Internet videos are not suitable for camera pose estimation.
They may be cartoons, contain substantial post-processing, or lack a clear frame of reference. 
Next, estimable videos are often low quality or do not contain scene dynamics.
Furthermore, even if suitable target videos can be identified, the process of estimating dynamic camera poses remains challenging.
Techniques that work well in static settings, such as structure-from-motion,  struggle with moving objects, varying appearances, and other complex dynamics common in Internet videos.

Given the challenges, most existing datasets take a different approach.
One line of work uses synthetic data~\cite{qiu2022airdos,butler2012naturalistic,mehl2023spring,zheng2023pointodyssey,mayer2016large,greff2022kubric} where ground-truth camera data can be obtained.
However, this often results in smaller-scale datasets (typically $<$500 videos) due to the high cost of 3D assets, and methods face a sim2real gap. 
Recent work finds camera data from real videos in restricted domains,
such as ego-centric videos~\cite{tschernezki2024epic}, self-driving videos~\cite{sun2020scalability}, hand-collected videos of pets~\cite{sinha2023common} or specific actions~\cite{grauman2023ego}.
These setups make SfM easier via dense viewpoints~\cite{tschernezki2024epic}, LiDAR sensory data~\cite{sun2020scalability}, multiple cameras~\cite{grauman2023ego,jin2024stereo4d} and turntable-style videos~\cite{sinha2023common}. However, such setups also impose significant constraints on the collected data.

We introduce \name, a large collection of dynamic Internet videos with camera annotations.
Sample videos and annotations appear in Figure~\ref{fig:teaser}. 
The dataset contains diverse content, a variety of dynamic object apparent sizes and targeted video length; and is analyzed in \S~\ref{sec:analysis}.
We address the central challenges of filtering for suitable dynamic video and annotating the corresponding camera poses with a carefully designed pipeline (\S~\ref{sec:curation}). 
Our filtering pipeline combines a mix of specialized experts that handle common reasons videos are unsuitable, as well as a generalist VLM that can detect and remove a variety of issues.
We produce accurate poses using a new method that merges the state-of-the-art in motion masking, correspondence tracking, and SfM. 
We apply this approach to Panda-70M~\cite{chen2024panda,xue2022advancing}, filtering 3.2M videos to produce 100K with high-quality camera information.

Our contributions are summarized as:

\begin{enumerate}[itemsep=0em]
    \item We introduce \name, a large-scale dynamic Internet video dataset annotated with camera information. Analysis shows the dataset is diverse in content and dynamics, and has targeted video length.
    \item We propose a filtering pipeline using specialist models and a VLM, motivated by Internet video analysis.
    \item We propose a dynamic pose estimation pipeline integrating state-of-the-art components in tracking and masking.
    \item Experiments show filtering selects videos with far higher precision than alternatives and pose estimation reduces error across metrics and settings by as much as 90\%.
\end{enumerate}

%% file: sections/2_related.tex
\section{Related Work}
\label{sec:related}

This work aims to collect and annotate dynamic Internet video with camera pose.
To do so, we face challenges in data curation and pose estimation.

\parnobf{Dynamic camera pose datasets} Due to challenges in predicting dynamic camera pose, datasets typically adopt strategies to replace or aid standard SfM.
Synthetic datasets are appealing as ground truth pose can be obtained~\cite{butler2012naturalistic,qiu2022airdos,zheng2023pointodyssey,mehl2023spring,zheng2023pointodyssey,greff2022kubric,mayer2016large}.
However, the requirement for engineered dynamic assets means the datasets are of small size and often contain limited dynamics \eg objects flying~\cite{mayer2016large} or falling~\cite{greff2022kubric}.
In real-world video, SfM is made easier with \eg turntable-style captures~\cite{sinha2023common}, dense viewpoints~\cite{tschernezki2024epic}, multiple cameras~\cite{grauman2023ego}, fisheye captures~\cite{jin2024stereo4d}, and LiDAR~\cite{sun2020scalability}.
We leverage the diverse Internet video dataset from Panda-70M~\cite{chen2024panda} and show our method can annotate high quality poses without relying on the constraints of prior work.
Concurrent works CamCo~\cite{xu2024camco} and B-Timer~\cite{liang2024btimer} collect 12K and 40K diverse videos~\cite{bain2021frozen}, respectively; we collect 100K and publicly release data. 

\input{figures/panda70m_stats}

\parnobf{Camera pose estimation}
Given many overlapping views of a static scene~\cite{ling2024dl3dv,zhou2018stereo,baruch1arkitscenes,schops2017multi,chang2017matterport3d,knapitsch2017tanks}, classical SfM~\cite{schonberger2016structure,snavely08,theia-manual} and SLAM~\cite{mur2015orb} are gold standards for precision, while learned methods recently are becoming competitive~\cite{smith2024flowmap,wang2023visual,wang2021tartanvo,teed2022deep,leroy2024mast3rsfm,yin2018geonet,zhang2022structure,kopf2021robust,teed2018deepv2d,cut3r} given large datasets~\cite{wang2020tartanair,reizenstein2021common,yeshwanth2023scannet++,zhou2018stereo}.
Dynamic Internet video is more challenging as dynamic correspondences cannot be used for bundle adjustment and also limit static correspondence.
Recent works~\cite{zhao2022particlesfm,li2021neural,liu2023robust,goli2024romo} mask dynamics using learned prediction~\cite{zhao2022particlesfm,chen2024leap,shen2023dytanvo} or motion and semantic cues~\cite{he2017mask,teed2020raft,hartley2003multiple}.
We upgrade masking using state-of-the-art methods in semantics, interaction, motion and tracking~\cite{jain2023oneformer,teed2020raft,hartley2003multiple,cheng2023putting,ravi2024sam2}.

Correspondence estimation in Internet video is also challenging due to varying lighting and appearance~\cite{li2021neural}.
As a result, we show masking+classical SfM~\cite{schonberger2016structure} fails on diverse Internet video.
ParticleSfM~\cite{zhao2022particlesfm} improves by inducing dense correspondences via propagated optical flow~\cite{teed2020raft} and uses more robust Global Bundle Adjustment, Theia-SfM~\cite{theia-manual}.
We adopt Theia-SfM and upgrade correspondence estimation by utilizing advances in long-term tracking~\cite{wang2023tracking,doersch2023tapir,harley2022particle,seidenschwarz2024dynomo} via BootsTAP~\cite{doersch2024bootstap}. 
Concurrent work TracksTo4D~\cite{kasten2024learning} predicts pose and dynamic masks given 2D tracks~\cite{karaev2023cotracker} but focuses on small dynamics and circular camera motion~\cite{sinha2023common}; we build off stronger 2D tracking BootsTAP and tackle diverse Internet video.
Concurrent work DATAP-SfM~\cite{Ye2024DATAP} trains joint tracking and motion prediction on synthetic FlyingThings3D~\cite{mayer2016large}, we instead leverage tracking and masking methods trained on large amounts of real data~\cite{doersch2024bootstap,ravi2024sam2,cheng2023towards,lin2014microsoft} and collect a large-scale dataset.
Concurrent works MonST3R~\cite{zhang2024monst3r} and MegaSaM~\cite{li2024_MegaSaM} generalize learned DUSt3R~\cite{wang2023dust3r} and DROID-SLAM~\cite{teed2021droid}, respectively, to dynamic scenes; we build around classical SfM for pose annotation given its precision in longer video.

%% file: figures/panda70m_stats.tex
\begin{figure*}[t]
    \centering
    \includegraphics[width=\linewidth]{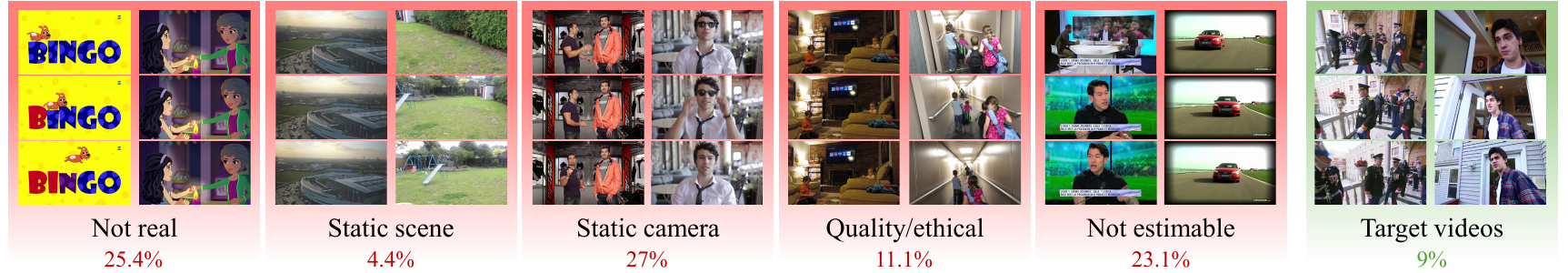}
    \caption{\textbf{Panda-Test dataset statistics}.
    Statistics reflect human labels on held-out 1K video Panda-Test set, detailed in \S~\ref{sec:analysis_panda70m_filtering}.
    Only 9\% are target dynamic camera pose estimation videos due to various issues, \eg static scene, low-quality or non-real content, and ambiguous or blurry frame of reference.
    We focus on moving cameras to facilitate downstream tasks \eg camera-controlled video generation and learned pose estimation.
    We remove unsuitable videos using a combination of specialist models and a generalist VLM.
    }
    \label{fig:panda70m_stats}
\end{figure*}

%% file: sections/3_method.tex
\section{\name: Dataset Curation}
\label{sec:curation}

\name is a large-scale video dataset comprising diverse dynamic Internet content with camera annotations.
Assembling such a dataset is challenging due to two primary issues related to camera pose estimation: (a) the vast majority of Internet videos are unsuitable for successful camera pose estimation, and (b) camera pose estimation on dynamic videos is inherently a challenging problem.
We address these challenges in this section.

\subsection{Candidate Selection Criteria}
\label{sec:curation_selection_criteria}

Most videos in representative Internet video datasets such as Panda-70M ~\cite{chen2024panda} are unsuitable for estimating camera pose: Figure~\ref{fig:panda70m_stats} shows only 9\% of videos from a randomly selected test set of 1,000 videos from Panda-70M dataset satisfy the criteria for camera pose estimation in dynamic settings.
We start by defining specific criteria to identify videos suitable for camera pose estimation, which we detect via subsequently defined data filtering steps.

We argue that videos should satisfy three criteria:
\begin{enumerate}[leftmargin=1.5em,itemsep=0.1em]
    \item[C1.] \textbf{Real-world and quality video.}
    Videos must be from the real world \ie not cartoons, computer screen recordings, side-by-side/composited videos, and heavily edited processed content.
    The videos should also be of sufficient quality, resolution, frame rate and should not have non-perspective distortion.
    \item[C2.] \textbf{Feasibility for pose prediction.}
    Videos should not contain severe zoom-in or zoom-out effects, abrupt shot changes, or ambiguous reference frames (\eg scenes filmed inside a moving car).
    Videos without static correspondences, \eg, when the background is fully blurred or occluded should be excluded.
    \item[C3.] \textbf{Dynamic camera and scene.}
    Non-static camera allows non-trivial pose for \eg training pose estimation or camera-controlled video generation methods.
    Videos with moving cameras and dynamic scenes typically also provide richer data, \eg, interactions involving people, supporting a wider range of downstream applications.
\end{enumerate}

\subsection{Candidate Video Selection}
\label{sec:curation_filtering}

The filtering process aims to automatically select videos matching these three criteria.
Our filtering pipeline identifies common reasons for unsuitability  using specialized models (\eg trained to detect distortion or estimate the camera focal length).
On the other hand, a variety of other problems, \eg, post-edited text are not effectively handled by these specialized models.
To this end, we use the recent vision-language models (VLM)~\cite{openai2024gpt4o} and prompt them to detect a variety of issues.
As will be shown in \S~\ref{sec:analysis_panda70m_filtering}, while individual filters (including the VLM) are insufficient on their own, their combined use is effective.

\parnobf{Filtering using task-specific models}
Specialized models are designed to address specific, frequently occurring reasons for video rejection.
These models are operated on temporally subsampled frames to enable efficient processing.

\begin{enumerate}[itemsep=0.00em]
    \item \emph{Cartoons and presenting}.
    This uses a classifier~\cite{cheng2023towards} to remove criteria in (C1) \eg cartoon or screen recording; and (C3) \eg static scenes of person sitting in front of camera or unlikely to interact.
    \item \emph{Non-perspective distortion}.
    Videos with high predicted distortion are removed.
    Distortion complicates applications that require reliable correspondences, so we consider them low quality (C1).
    \item \emph{Focal length}.
    Videos with high predicted focal length variance are removed, as they often contain zoom effects or shot changes (C2).
    We also exclude videos with long focal lengths, which typically include backgrounds that are too blurred for reliable pose estimation (C2) or are sporting shots with static cameras (C3).
    \item \emph{Dynamic object masking}.
    Videos with excessively large predicted mask sizes are considered unsuitable for pose estimation (C2) and are removed, as reliable pose estimation requires sufficient static correspondences. 
    \item \emph{Optical flow}.
    High peak sequential optical flow is typically due to shot change (C2); such videos are removed using predicted flow.
    Low mean sequential flow typically indicates static videos (C3); these are also removed. 
    \item \emph{Point tracking}.
    Abrupt disappearance of predicted point tracks can indicate shot change, blurred background, or severe zoom-in/zoom-out, making the video unsuitable for pose estimation (C2).
    Conversely, extremely stable tracks suggest neither the camera nor the scene has moved significantly (C3). Both cases are removed.
\end{enumerate}

\parnobf{Filtering using general VLM}
The generalist VLM addresses a variety of possibilities for video rejection.
We prompt GPT-4o mini~\cite{openai2024gpt4o,openai2023gpt4} a series of eight questions addressing all three classes of criteria.
They cover: if the video is taken from an ambiguous reference, if the background is too blurry to match correspondence, if frames are distorted or have very long focal length, if the scene is static, if the video is a cartoon, has been post-processed, and if children are present.
We find having overlap with specialist filters, \eg identifying long focal length, is helpful, since the VLM can be prompted to consider diverse cues for removal.

\input{figures/annotation_approach}

\vspace{0mm}

\parnobf{Collection details}
Beginning from 3.2M Panda-70M~\cite{chen2024panda} videos, we apply filtering (\S~\ref{sec:curation_filtering}) leaving 137K dynamic videos.
We estimate pose on 107K and drop trajectories with less than 80\% of frames registered, leaving 100k videos.
We apply filters sequentially for efficiency.
First are lightweight Hands23, optical flow and focal length filters on the full set.
We remove low scoring videos, leaving 1.63M videos.
We next run the distort filter and remove low scores, leaving 1.53M;
then tracking, leaving 679K; then masking, leaving 462K; and finally the VLM, after which we apply a strict final filter to get 137K for pose estimation.

\subsection{Dynamic Camera Pose Estimation}
\label{sec:curation_pose}

Having collected suitable videos, our objective is to obtain high-quality camera poses.
Pose estimation on dynamic Internet video is challenging: not only do dynamic objects occlude the underlying static scene,
the static scene can change appearance making correspondence estimation difficult.
Our approach addresses both challenges (Figure~\ref{fig:annotation_approach}).
For masking, we use state-of-the-art methods in semantics, interaction, motion, and tracking, with each component improving overall performance. 
For correspondences, we use the latest point tracking method~\cite{doersch2024bootstap} applied within a sliding window on the video.
Finally, we apply global bundle adjustment~\cite{theia-manual}, which has shown to be effective in handling the challenges posed by Internet videos~\cite{zhao2022particlesfm}.

\parnobf{Dynamic masking}
We first aim to segment {\it dynamic} regions in an input video.
Our approach uses advances in several communities to produce accurate masks.

\begin{enumerate}[itemsep=-0.1em]
    \item \emph{Semantic segmentation}.
    We segment common dynamic classes (\eg humans, vehicles, animals, and sports equipment) using OneFormer~\cite{jain2023oneformer}.
    \item \emph{Object interaction segmentation}.
    This masks held objects which may dynamic yet outside semantic classes \eg silverware.
    We use the Hands23~\cite{cheng2023towards} hand-object interaction segmentation model.
    \item \emph{Motion segmentation}.
    This handles dynamic objects that are not common classes or involve human manipulation, \eg rustling leaves or flowing water.
    We use motion masking of RoDynRF~\cite{liu2023robust}, which removes regions with high Sampson error~\cite{hartley2003multiple} based on optical flow~\cite{teed2020raft}.
    \item \emph{Mask propagation}.
    Propagation provides smooth masks across frames and precise object boundaries.
    We use SAM2~\cite{ravi2024sam2} to propagate masks.
\end{enumerate}

\parnobf{Point tracking}
We estimate correspondences using point tracking, which uses temporal information in videos to improve estimation.
Unlike pairwise estimation, tracking takes advantage of the fact that points generally move only small amounts between sequential frames.
Additionally, tracking the same point over multiple frames yields rapidly increasing paired correspondences across each frame combination.
For this purpose, we use BootsTAP~\cite{doersch2024bootstap} to track a grid of points forward several frames, move forward, and repeat in sliding window fashion.
Grid-based point tracking facilitates denser correspondences, while the extended tracking duration supports long-term correspondences, reducing drift and aiding in loop closure.
Sliding window tracking of point grids ensures that each frame maintains a sufficient number of correspondences in the case previous windows have become occluded or faced drift.

\parnobf{Global bundle adjustment}
We use global bundle adjustment method Theia-SfM~\cite{theia-manual}, using correspondences from tracklets as input.
From a single tracklet, we extract correspondences for all pairs, excluding pairs containing a frame in which the tracklet is contained within a dynamic mask.

\input{tables/dataset_analysis}

%% file: figures/annotation_approach.tex
\begin{figure}[t]
    \centering
    \includegraphics[width=\linewidth]{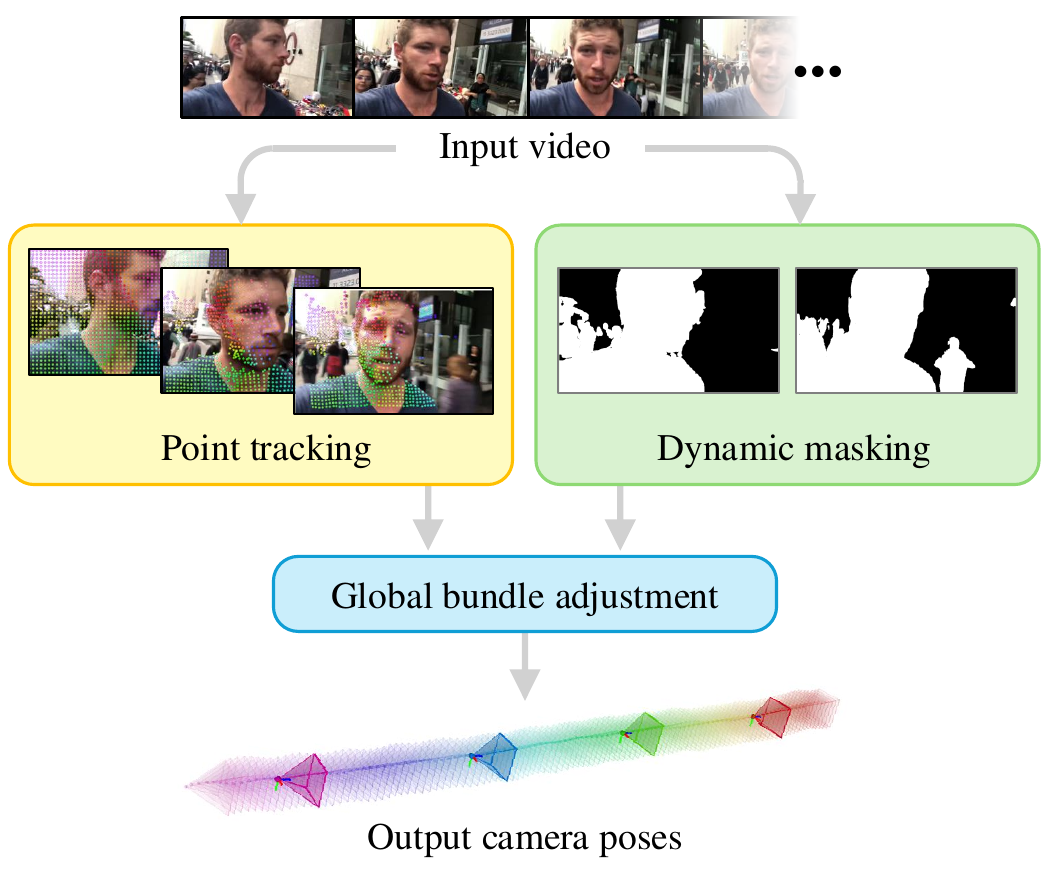}
    \caption{\textbf{Pose estimation approach}.
    We apply the state-of-the-art point tracking method at a sliding window to produce dense, long-term correspondences.
    Complementary dynamic masks are used to remove non-static tracks.
    The remaining static tracks are provided as input to global bundle adjustment.
    }
    \label{fig:annotation_approach}
\end{figure}

%% file: tables/dataset_analysis.tex
\begin{table*}[t]
\captionsetup{type=figure}
\makebox[0pt][l]{%
\begin{tabular}{@{} c @{\extracolsep{10pt}} c @{}}
    \begin{minipage}{0.58\textwidth}
        \resizebox{\ifdim\width>\linewidth \linewidth \else \width \fi}{!}{
            \begin{tabular}{lccccc}
                \toprule
                Dataset & Real/Syn. & Num. vids. & Num. frames & Domain & Access \\
                \midrule
                T.Air Shibuya~\cite{qiu2022airdos} & Syn. & 7 & 0.7K & Street & \textbf{Public} \\
                MPI Sintel~\cite{butler2012naturalistic} &  Syn. & 14 & 0.7K & Scripted & \textbf{Public} \\
                PointOdyssey~\cite{zheng2023pointodyssey} & Syn. & 131 & 200K & Walking & \textbf{Public} \\
                FlyingThings3D~\cite{mayer2016large} & Syn. & 220 & 2K & Objects & \textbf{Public} \\
                Kubric Movi-E~\cite{greff2022kubric} & Syn. & 400 & 10K & Objects & \textbf{Public} \\
                EpicFields~\cite{tschernezki2024epic} & Real & 671 & \underline{19,000K} & Kitchens & \textbf{Public} \\
                Waymo~\cite{sun2020scalability} & Real & 1,150 & 200K & Driving & \textbf{Public} \\
                CoP3D~\cite{sinha2023common} & Real & 4,200 & 600K & Pets & \textbf{Public} \\
                Ego-Exo4D~\cite{grauman2023ego} & Real & 5,035 & \textbf{23,000K} & Set tasks & \textbf{Public} \\
                Stereo4D~\cite{jin2024stereo4d} & Real & \textbf{110,000} & 10,000K & S. fisheye & \textbf{Public} \\ 
                CamCo~\cite{xu2024camco} & Real & 12,000 & 385K & \textbf{Diverse} & Private \\
                B-Timer~\cite{liang2024btimer} & Real & 40,000 & \underline{19,000K} & \textbf{Diverse} & Private \\
                \name & Real & \underline{100,131} & 6,806K & \textbf{Diverse} & \textbf{Public} \\
                \bottomrule
            \end{tabular}
        }
        \captionof{table}{\textbf{Dynamic camera pose datasets.}
        \name has the most videos of diverse Internet video datasets.
        Datasets with more frames are private or more uniform; \eg stereo fisheye is typically outdoor PoV walking.
        We use short videos, yielding fewer frames but high dynamics (Figure~\ref{fig:length}).
        }
        \label{tab:size}
    \end{minipage}
    &
    \begin{minipage}{0.40\textwidth}
        \begin{minipage}{\textwidth}
            \includegraphics[width=\textwidth]{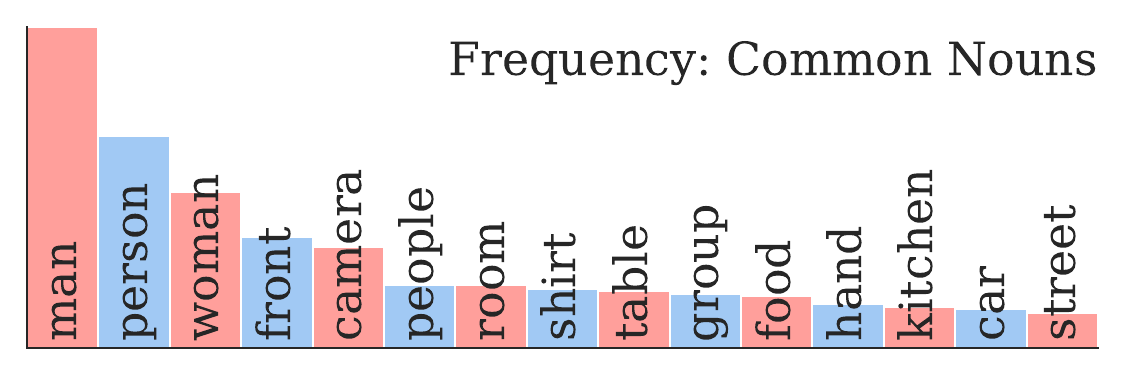}
        \end{minipage}
        \begin{minipage}{\textwidth}
            \includegraphics[width=\textwidth]{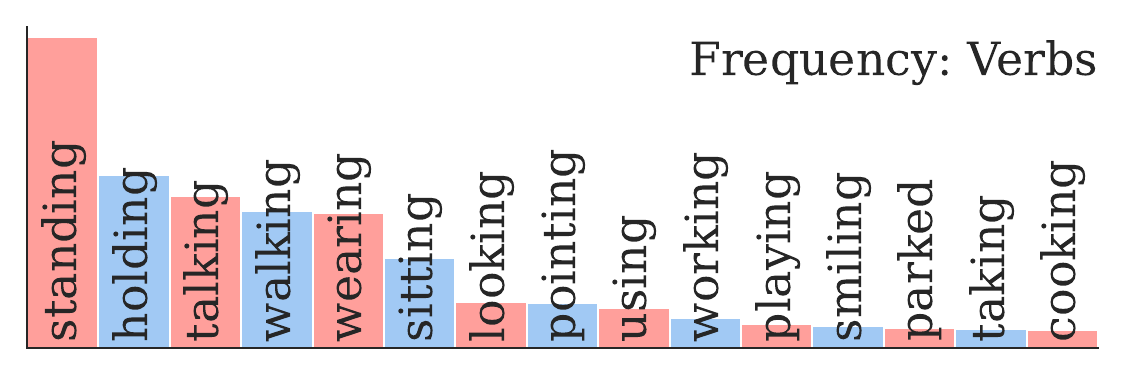}
        \end{minipage}
        \captionof{figure}{\textbf{Diverse content.} Frequent nouns cover varied subjects: person, hand, car; objects: shirt, table, food; and settings: room, kitchen, street. Verbs span diverse actions: using, working, playing.}
        \label{fig:caption}
    \end{minipage}
\end{tabular}
}
\end{table*}

%% file: sections/4_datasets.tex
\section{\name: Dataset Analysis}
\label{sec:analysis}

In this work, we focus on dynamic Internet videos due to their large scale and diversity.
We evaluate the effectiveness of our filtering pipeline in identifying videos suitable for pose estimation (\S~\ref{sec:analysis_panda70m_filtering}) and provide a statistical analysis of the resulting dataset (\S~\ref{sec:analysis_stats}).

\subsection{Filtering Evaluation on Panda-Test}
\label{sec:analysis_panda70m_filtering}

Our goal for filtering is to identify suitable videos for pose estimation given diverse Internet video.

\parnobf{Dataset}
We evaluate filtering performance on 1,000 randomly selected held-out videos from Panda-70M, which we refer to as Panda-Test.
Each video is manually classified as suitable or unsuitable for pose estimation, yielding 90 (9\%) suitable videos.
These 90 videos exhibit similar diversity to \name, \eg content, lengths, and dynamic sizes.

\parnobf{Metrics}
Our primary evaluation metrics are precision and recall.
We prioritize precision, aiming to minimize the number of inaccurate videos included in \name.
However, recall is also important, as a low recall would mean a larger number of necessary videos to process to achieve a dataset of the same size.

\input{tables/panda70m_filtering}

\parnobf{Baselines and ablations}
We compare our method with existing and alternative filtering methods.
We filter using Our SfM reconstructed points (CamCo~\cite{xu2024camco}) and reprojection error (inspired by B-Timer~\cite{liang2024btimer}), along with video suitability and interaction classifier Hands23~\cite{cheng2023towards}.
In addition, we prompt GPT-4o mini~\cite{openai2024gpt4o} in a manner similar to our filtering pipeline (\S~\ref{sec:curation_filtering}).
We implement two versions of this comparison: one producing binary outputs and another assigning scores to each example.
Finally, we conduct ablation studies on each step of our filtering method from \S~\ref{sec:curation_filtering}.

\parnobf{Results}
Figure~\ref{fig:panda_filter} shows our filtering selects videos with high precision and recall, outperforming baseline methods.
In fact, filtering has test precision of 0.78 at the threshold used for collecting \name, a level that no baseline achieves at any recall, except for reprojection error at 0.02 recall.
Each component in our method improves performance, with the VLM providing a large boost, even though it is relatively weak as a standalone approach.

\subsection{Dataset statistics}
\label{sec:analysis_stats}

Having collected a substantial number of target videos, we assess the characteristics of the resulting dataset.
In addition to evaluating the dataset size, we examine whether resulting videos exhibit desirable attributes, \eg diverse content, appropriate video length and diverse dynamic object size.

\parnobf{Dataset size}
Table~\ref{tab:size} shows that \name\ contains many more videos than existing diverse datasets.
The largest alternatives are typically limited to specific settings, such as kitchens, driving, walking, or pet turntable videos~\cite{jin2024stereo4d,sinha2023common,grauman2023ego,sun2020scalability,tschernezki2024epic}.
Concurrent works CamCo~\cite{xu2024camco} and B-Timer~\cite{liang2024btimer} also feature diverse content but have fewer videos and are private.

\parnobf{Video content} Figure~\ref{fig:caption} shows frequent nouns and verbs that appear in Panda-70M's~\cite{chen2024panda} captions associated with \name videos.
Common nouns cover a wide range of subjects, objects and settings; while verbs reflect a variety of actions.
This linguistic variety provides evidence \name contains videos with diverse content.

\parnobf{Video length} Figure~\ref{fig:length}, left shows the distribution of video lengths in \name.
The majority of videos span 4 to 10 seconds.
These short clips are typically rich in dynamics, while being long enough for substantial camera motion.

\input{tables/dataset_analysis2}

\input{tables/lightspeed}
\input{figures/qualitative_lightspeed_v2}

\parnobf{Dynamic object size} Figure~\ref{fig:length}, right shows the distribution of apparent dynamic object sizes in \name, ranging from small to large.
Large dynamic objects occlude static correspondences, making pose estimation challenging.
Smaller apparent dynamic objects typically correspond to objects further away, which can move quickly, making precise masking challenging.  
Videos where dynamic objects occupy nearly the entire frame are filtered out as they make pose estimation infeasible.

%% file: tables/panda70m_filtering.tex
\begin{figure}
    \centering
    \includegraphics[height=1.65in]{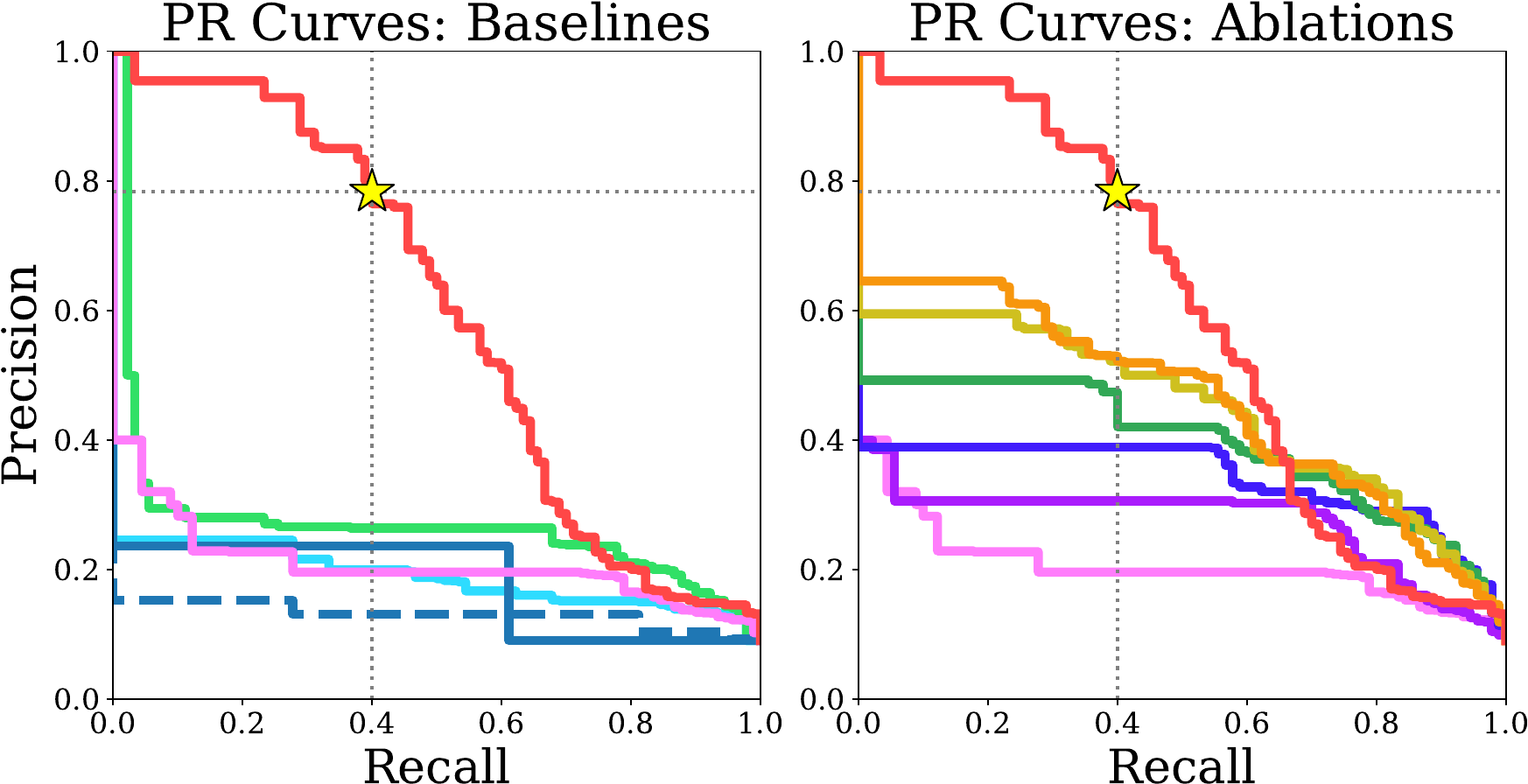}
    \caption{
        \textbf{Dynamic video filtering on Panda-Test.} We show PR curves for baselines and ablations. Our filtering surpasses all baselines and ablations by a considerable margin.
        The\protect\yellowstar represents \name's operating thresholds.
        For baselines, we show:
        \textcolor{recon_pts}{$\blacksquare$} Reconstructed points (CamCo~\cite{xu2024camco}),
        \textcolor{reproj_err}{$\blacksquare$} Reprojection error,
        (\textcolor{vlm}{solid $\blacksquare$}) GPT-4o mini~\cite{openai2024gpt4o}: binary,
        (\textcolor{vlm}{dashed $\blacksquare$}) GPT-4o mini~\cite{openai2024gpt4o}: score,
        \textcolor{hands23}{$\blacksquare$} Hands23~\cite{cheng2023towards}, and
        \textcolor{dynpose}{$\blacksquare$} Ours.
        For ablations, we begin from \textcolor{hands23}{$\blacksquare$} Hands23 and add components until we recover \textcolor{dynpose}{$\blacksquare$} Ours. Specifically, we depict:
        \textcolor{hands23}{$\blacksquare$} Hands23,
        \textcolor{flow}{$\blacksquare$} +Flow,
        \textcolor{tracking}{$\blacksquare$} +Tracking, 
        \textcolor{masking}{$\blacksquare$} +Masking,
        \textcolor{focal}{$\blacksquare$} +Focal,
        \textcolor{distort}{$\blacksquare$} +Distort,
        \textcolor{dynpose}{$\blacksquare$} +VLM (Ours).
    }
    \label{fig:panda_filter}
\end{figure}

%% file: tables/dataset_analysis2.tex
\begin{table}[t]
\captionsetup{type=figure}
\resizebox{\ifdim\width>\columnwidth \columnwidth \else \width \fi}{!}{
\begin{tabular}{@{} c @{\extracolsep{5pt}} c @{}}
    \begin{minipage}{0.47\textwidth}
        \includegraphics[width=\linewidth]{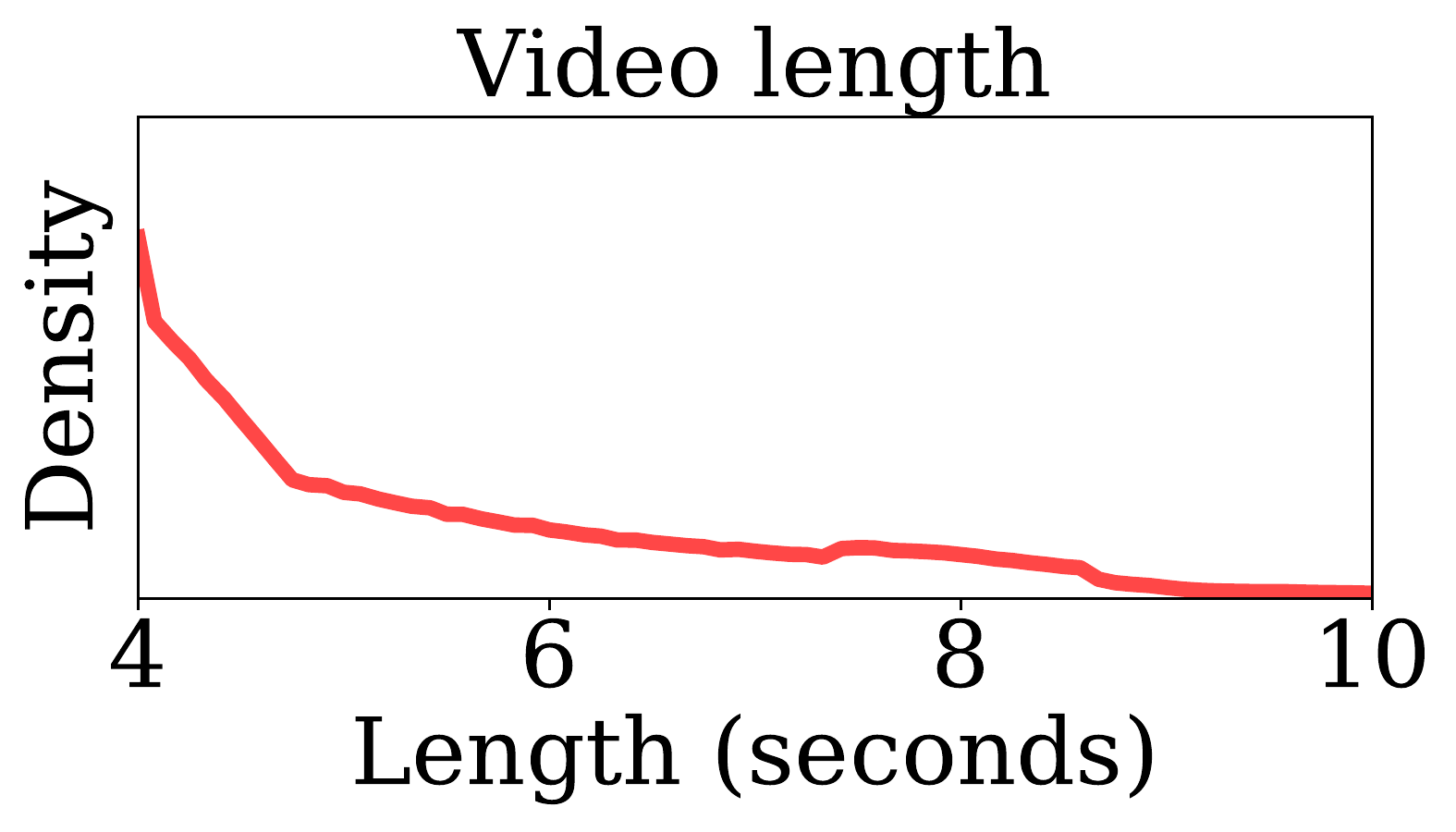}
    \end{minipage} 
    &
    \begin{minipage}{0.47\textwidth}
        \includegraphics[width=\linewidth]{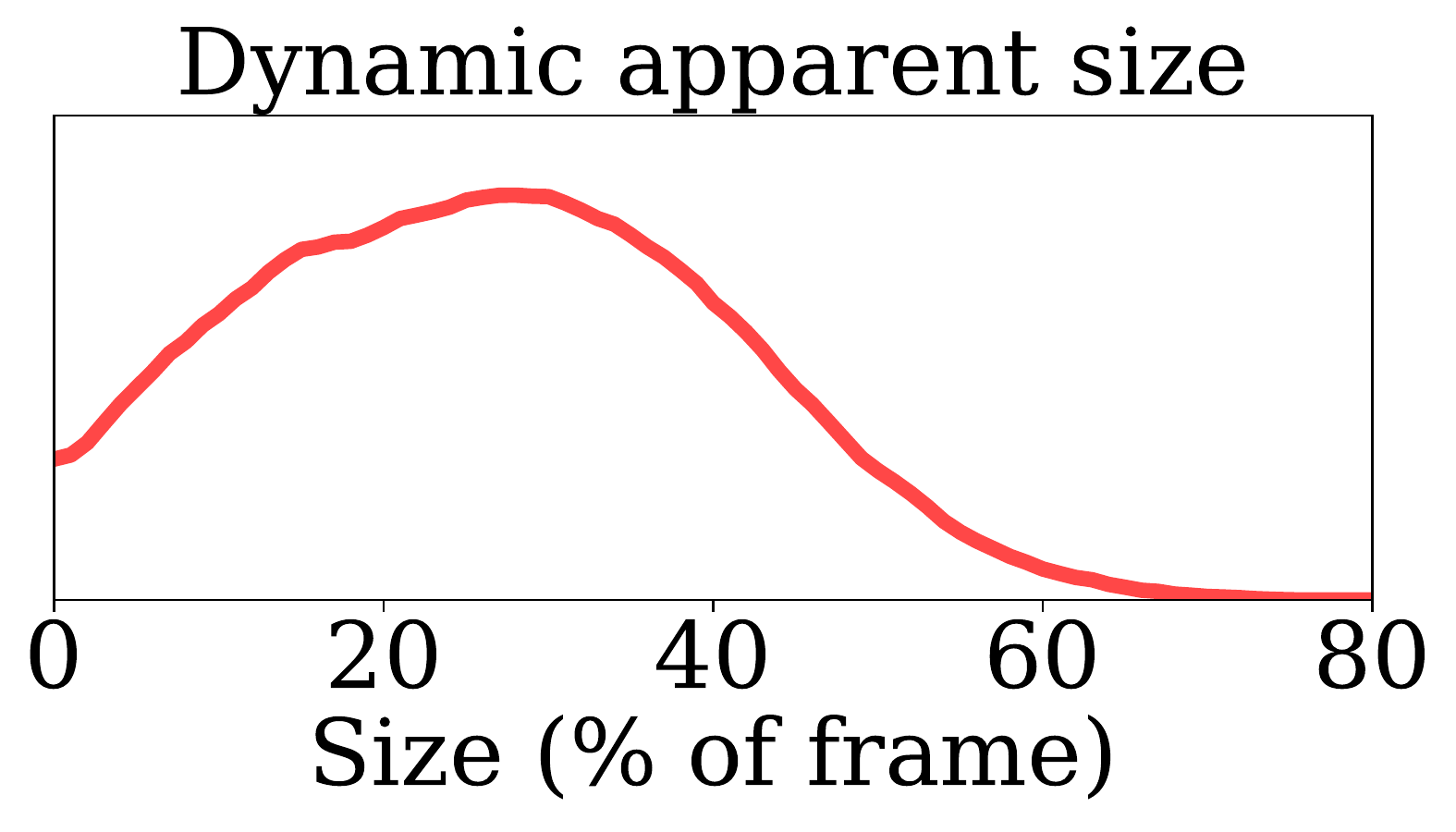} 
    \end{minipage}
\end{tabular}
}
\captionof{figure}{\textbf{Left: Targeted video length.} \name videos are primarily 4-10s, ideal for dynamic pose: shorter videos contain little ego-motion, longer videos have less dense dynamics and ego-motion.
\textbf{Right: Diverse dynamic apparent size.} Mean size in \% across video. Large dynamic objects occlude static correspondences, making pose estimation challenging. Videos may average small size in the case of only a few dynamic frames.}
\label{fig:length}
\end{table}

%% file: tables/lightspeed.tex
\begin{table*}
\centering
\resizebox{\ifdim\width>\linewidth \linewidth \else \width \fi}{!}{

\begin{tabular}{l c c c c c c c c} \toprule
  & \multicolumn{4}{c}{All 36 videos (Iden. Rot. + Rand. Tr. if fail)} & \multicolumn{3}{c}{8 videos: all succeed only} \\
Method & \% Vids. reg.$\uparrow$ & ATE (m)$\downarrow$ & RPE Tr. (m)$\downarrow$ & RPE Rot. ($^{\circ}$)$\downarrow$ & ATE (m)$\downarrow$ & RPE Tr. (m)$\downarrow$ & RPE Rot. ($^{\circ}$)$\downarrow$ \\
\midrule
Iden. Rot. + Rand. Tr. & \textbf{100.} & 0.652 & 0.139 & 1.60 & 0.390 & 0.080 & 1.00 \\
DROID-SLAM~\cite{teed2021droid} & \textbf{100.} & 0.198 & \underline{0.046} & 1.75 & 0.048 & 0.017 & 0.82 \\
DUSt3R~\cite{wang2023dust3r} & 97.2 & 0.412 & 0.177 & 20.1 & 0.256 & 0.124 & 18.5   \\
MonST3R~\cite{zhang2024monst3r} & \textbf{100.} & \underline{0.149} & \underline{0.046} & \textbf{1.21} & \underline{0.036} & \underline{0.011} & \underline{0.46} &  \\
LEAP-VO~\cite{chen2024leap} & \textbf{100.} & 0.206 & 0.049 & 1.70 & 0.037 & \underline{0.011} & 0.73 \\
COLMAP~\cite{schonberger2016structure} & 44.4 & 0.388 & 0.082 & 2.03 & 0.122 & 0.026 & 1.91  \\
COLMAP+Mask~\cite{schonberger2016structure} & 38.9 & 0.323 & 0.085 & 1.64 & 0.089 & 0.017 & 1.36 \\
ParticleSfM~\cite{zhao2022particlesfm} & 97.2 & 0.185 & 0.075 & 2.99 & 0.051 & 0.047 & 2.91 \\
Ours & \textbf{100.} & \textbf{0.072} & \textbf{0.033} & \underline{1.31} & \textbf{0.003} & \textbf{0.002} & \textbf{0.30} \\
\bottomrule
\end{tabular}
}
    \caption{\textbf{Camera pose estimation on Lightspeed.} Our pose estimation algorithm registers all sequences and cuts trajectory error by 50\% across all sequences (left) and 90\% on easy sequences (right) \vs all other methods.
    }
\label{tab:lightspeed}
\end{table*}

%% file: figures/qualitative_lightspeed_v2.tex
\begin{figure*}[t]
    \centering
    \includegraphics[width=\linewidth]{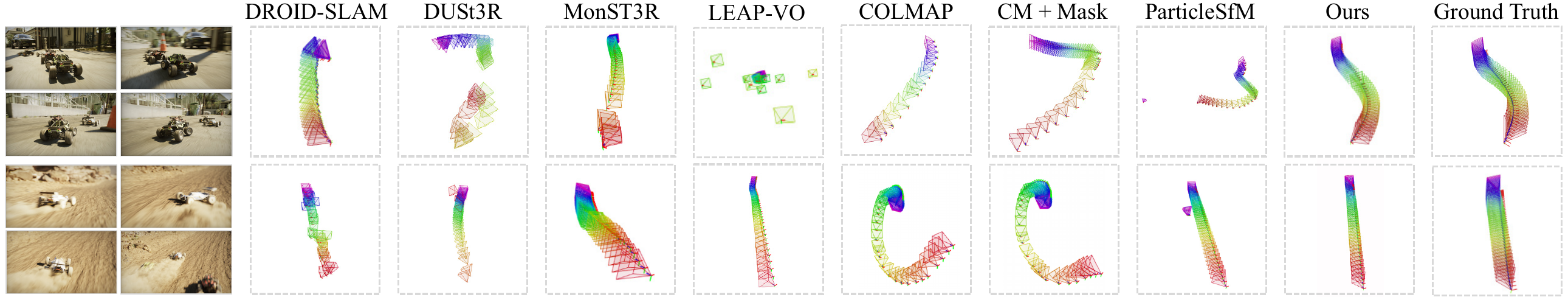}
    \caption{\textbf{Predicted trajectories on Lightspeed}. Pose sequence over time: \textcolor{red}{R}\textcolor{orange}{O}\textcolor{yellow}{Y}\textcolor{green}{G}\textcolor{blue}{B}\textcolor{violet}{V}.
    We visualize photorealistic renderings of Lightspeed left.
    Static methods DROID-SLAM, COLMAP and DUSt3R struggle in this dynamic setting, failing to register a consistent sequence or putting too much or too little curvature. 
    Top: dynamic methods MonST3R, LEAP-VO and ParticleSfM do not produce smooth sequences.
    Bottom: MonST3R, LEAP-VO and COLMAP+Mask add curvature.
    Ours produces smooth and accurate trajectories.
    }
    \label{fig:qualitative_lightspeed}
\end{figure*}

%% file: sections/5_experiments.tex
\section{Evaluation on Camera Pose Estimation}
\label{sec:experiments}

We proceed to evaluate the pose estimation efficacy of the curated dataset \name through controlled experiments.
Direct evaluation on dynamic Internet videos is challenging due to the lack of ground truth camera pose data.
Our evaluation approach is therefore two-fold: (1) we use a photorealistic synthetic rendering dataset, Lightspeed, which provides ground truth poses for direct comparison; (2) we evaluate directly on Internet videos by annotating 10K precise correspondences on Panda-Test, enabling pose evaluation using Sampson error~\cite{hartley2003multiple}.

\input{tables/panda70m}
\input{figures/qualitative_panda}

\parnobf{Baselines and ablations}
Our primary point of comparison throughout is ParticleSfM~\cite{zhao2022particlesfm}, which shares our general pipeline involving static tracking, dynamic masking, and global SfM.
We also compare against standard SfM method COLMAP~\cite{schonberger2016structure}, both in its original form and with correspondences filtered using dynamic masking.
Finally, we evaluate our method against state-of-the-art learning-based static methods DROID-SLAM~\cite{teed2021droid} and DUSt3R~\cite{wang2023dust3r}, and dynamic methods LEAP-VO~\cite{chen2024leap} and MonST3R~\cite{zhang2024monst3r}.

\subsection{Pose Evaluation on Lightspeed}
\label{sec:experiments_lightspeed}

We introduce Lightspeed, a challenging, photorealistic benchmark for dynamic pose estimation containing ground truth camera pose.
This dataset is a good benchmark for pose estimation because it is shares several important characteristics of \name: diverse environments, large dynamic objects and varied clip lengths of several seconds.

\parnobf{Dataset} We use NVIDIA Racer RTX~\cite{nvidia_racer_rtx}, shown in Figure~\ref{fig:qualitative_lightspeed}, which follows several RC cars moving quickly through challenging indoor and outdoor scenes, featuring day and night lighting and first and third person views.
We split the video into clips, removing ones with non-perspective distortion, blurry background, static scenes, or cameras with static or have trivial linear trajectory (following~\cite{Wulff:ECCVws:2012}), resulting in 36 sequences. We extract ground truth camera pose from corresponding 3D Assets.

\parnobf{Metrics} We follow prior work~\cite{liu2023robust,zhao2022particlesfm,shen2023dytanvo} in reporting Average Trajectory Error (ATE) as well as Relative Pose Error (RPE) for rotation and translation.
For full details see prior work. 
To summarize, metrics are computed after trajectory alignment and scaling, and relative error is computed on sequential frames.
To compute scores across all videos, we replace predicted trajectories that do not converge with random uniform translations and identity rotations.
We also separately report scores on the subset of videos upon which all methods converge.

\parnobf{Results} Table~\ref{tab:lightspeed} shows COLMAP and COLMAP+Mask struggle to register many challenging sequences in Lightspeed.
DROID-SLAM, DUSt3R, LEAP-VO and ParticleSfM provide registration, but are inaccurate.
MonST3R offers better trajectory error, though Ours is clearly superior, cutting trajectory error by 50\% (all videos) and 90\% (succeed only).
All alternatives struggle in at least one sequence in Figure~\ref{fig:qualitative_lightspeed}, while Ours handles both.

\subsection{Pose Evaluation on Panda-Test}
\label{sec:experiments_panda70m_pose}

Next, we evaluate on dynamic Internet videos.
We use the 90-video subset from Panda-Test (\S~\ref{sec:analysis_panda70m_filtering}), featuring various challenging objects with estimable camera poses.

\parnobf{Metrics}
Ground truth camera poses are unavailable in unlabeled dynamic Internet videos, so we opt to annotate correspondences and compute Sampson error from the predicted poses.
We curate 10K correspondences across unique image pairs spanning the 90 videos to use for evaluation.
Pairs are manually annotated by expert annotators to achieve coarse accuracy. 
These matches are refined using SuperPoint~\cite{detone2018superpoint} and LightGlue~\cite{lindenberger2023lightglue} by selecting the LightGlue match found within 10 pixels of the human-labeled points on 720p images.
If no such match is found, the correspondence is discarded.
Frame pairs are selected randomly from within 2.5 seconds of each other.

Reprojection error for each pair is the square root of Sampson error.
Sampson error computes the squared distance from an annotated correspondence to the epipolar line computed using the fundamental matrix based on predicted relative camera pose and intrinsics.
Metrics are averaged for all pairs in each video, enabling per-video analysis.
Our main metrics are mean error across videos, and percent of videos with mean reprojection error within a threshold; we use 5, 10 and 30 pixels at normalized 720p resolution.
Non-registered frames are filled with nearest neighbor poses.
If an entire sequence is not registered, the identity is used.

\input{tables/finetune}

\parnobf{Results}
Table~\ref{tab:panda} shows similar trends to Lightspeed: static methods DROID-SLAM and DUSt3R are inaccurate while COLMAP, even with dynamic masking, struggles to register many videos.
ParticleSfM improves, but contains high mean error.
MonST3R and LEAP-VO register all frames, but Ours reduces error across metrics. 
This includes more than 35\% reduction in mean error vs. all methods on the 52 sequences all methods succeed.
Figure~\ref{fig:qualitative_panda} shows qualitative results from two challenging sequences.
Qualitative results agree with quantitative: DROID-SLAM, DUSt3R and COLMAP struggle, while MonST3R, LEAP-VO, COLMAP+Mask and ParticleSfM face failure cases.
Ours better handles difficult cases \eg large dynamics or varied appearance.
Evidently, the combined state-of-the-art components used for masking and tracking (\S~\ref{sec:curation_pose}) yield a pipeline effective at each corresponding task.

%% file: tables/panda70m.tex
\begin{table*}
\resizebox{\ifdim\width>\linewidth \linewidth \else \width \fi}{!}{
\begin{tabular}{l c c c c c c c c c} \toprule
  & \multicolumn{5}{c}{All 90 estimable videos (Identity if fail)} & \multicolumn{4}{c}{52 videos: all succeed only} \\
 & \% Vids. & \multicolumn{4}{c}{Mean per-video reprojection error, 720p} & \multicolumn{4}{c}{Mean per-video reprojection error, 720p} \\
\textit{Baselines} & registered$\uparrow$ & $\%<5 $Pix$\uparrow$ & $\%<10$ Pix$\uparrow$ & $\%<30$ Pix$\uparrow$ & Mean$\downarrow$ & $\%<5$ Pix$\uparrow$ & $\%<10$ Pix$\uparrow$ & $\%<30$ Pix$\uparrow$ & Mean$\downarrow$ \\
\midrule
Identity & \textbf{100.} & 0.0 & 0.0 & 2.2 & 151 & 0.0 & 0.0 & 0.0 & 133 \\
DROID-SLAM~\cite{teed2021droid} & \textbf{100.} & 57.8 & 77.8 & 94.4 & 11.0 & 59.6 & 84.6 & 96.2 & 6.78  \\
DUSt3R~\cite{wang2023dust3r} & 96.7 & 0.0 & 6.7 & 48.9 & 43.0 & 0.0 & 9.6 & 57.7 & 30.3 \\
MonST3R~\cite{zhang2024monst3r} & \textbf{100.} & 55.6 & \underline{78.9} & 90.0 & 9.86 & 63.5 & 84.6 & 90.4 & 9.71 \\
LEAP-VO~\cite{chen2024leap} & \textbf{100.} & 64.4 & 76.7 & \underline{96.7} & \underline{7.59} & 75.0 & 84.6 & \underline{98.1} & \underline{6.03} \\
COLMAP~\cite{schonberger2016structure} & 82.2 & 51.1 & 62.2 & 73.3 & 27.5 & 71.2 & 82.7 & 92.3 & 9.03 \\
COLMAP+Mask~\cite{schonberger2016structure} & 67.8 & 47.8 & 58.9 & 75.6 & 30.1 & 69.2 & 82.7 & 96.2 & 6.10 \\
ParticleSfM~\cite{zhao2022particlesfm} & 92.2 & \underline{70.0} & 76.7 & 88.9 & 12.5 & \underline{80.8} & \underline{86.5} & 96.2 & 6.77 \\ 
Ours & 95.6 & \textbf{72.2} & \textbf{84.4} & \textbf{98.9} & \textbf{5.76} & \textbf{82.7} & \textbf{94.2} & \textbf{100.} & \textbf{3.75} \\ 
\bottomrule
\end{tabular}
}
\caption{\textbf{Camera pose estimation on Panda-Test.}
Reprojection error on 10K image pairs, by video, normalized to 720p. 
Static methods DUSt3R and COLMAP struggle faced with dynamics while DROID-SLAM lacks precision. 
ParticleSfM registers more videos than COLMAP+Mask but both fall short of Ours in registration and accuracy. MonST3R and LEAP-VO register all frames leading to moderate errors, but Ours outperforms across metrics.
}
\label{tab:panda}
\end{table*}

%% file: figures/qualitative_panda.tex
\begin{figure*}[t]
    \centering
    \includegraphics[width=\linewidth]{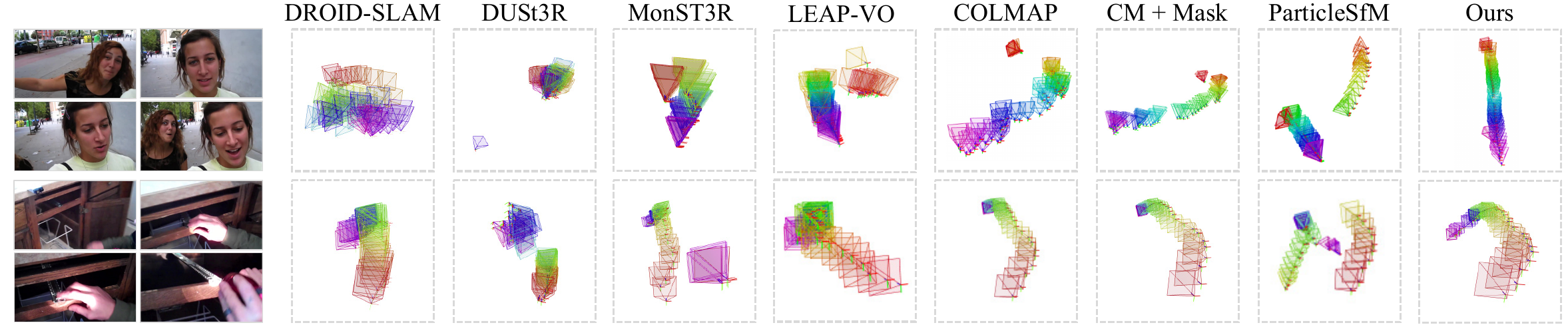}
    \caption{\textbf{Predicted trajectories on Panda-Test}.
    Pose sequence over time: \textcolor{red}{R}\textcolor{orange}{O}\textcolor{yellow}{Y}\textcolor{green}{G}\textcolor{blue}{B}\textcolor{violet}{V}.
    Static methods like DROID-SLAM, DUSt3R and COLMAP struggle faced with dynamics.
    MonST3R, LEAP-VO, COLMAP+Mask and ParticleSfM can struggle with large dynamic regions (top) and tracking across varied appearance and lighting (bottom); while Ours handles these cases.
    }
    \label{fig:qualitative_panda}
\end{figure*}

%% file: tables/finetune.tex
\begin{table}
\resizebox{\ifdim\width>\columnwidth \columnwidth \else \width \fi}{!}{
\begin{tabular}{l c c c c c}
  \toprule
& \multicolumn{5}{c}{All 90 estimable videos (Identity if fail)} \\
& \% Vids. & \multicolumn{4}{c}{Mean per-video reprojection error, 720p} \\
\textit{Method} & reg.$\uparrow$ & $\%<5 $Px$\uparrow$ & $\%<10$ Px$\uparrow$ & $\%<30$ Px$\uparrow$ & Mean$\downarrow$ \\
\midrule
 DUSt3R & 96.7 & 0.0 & 6.7 & 48.9 & 43.0 \\
+ Synthetic (MonST3R~\cite{zhang2024monst3r}) & \textbf{100.} & \textbf{55.6} & 78.9 & 90.0 & 9.86 \\
+ DynPose-100K (Ours) & \textbf{100.} & 54.4 & \textbf{82.2} & \textbf{92.2} & \textbf{8.78} \\
\bottomrule
\end{tabular}
}
\par\vspace{3mm}
\noindent\begin{minipage}{\linewidth}
    \centering
    \includegraphics[width=\linewidth]{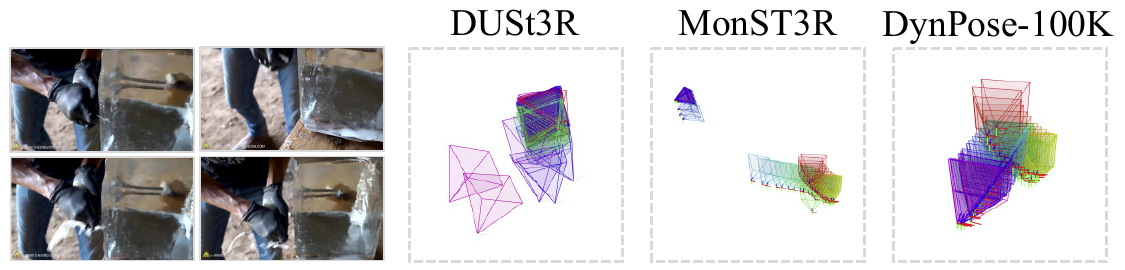}
\end{minipage}
\captionsetup{skip=9pt}
\captionof{table}{\textbf{Camera estimation on Panda-Test.}
DynPose-100K fine-tuning of DUSt3R has lower mean error and similar to or better than accuracy compared to synthetic data (MonST3R). We train with only 2K videos / 140K frames, smaller than the 1.3M frames used to train MonST3R; demonstrating efficient supervision.}
\label{tab:finetune}
\end{table}

%% file: sections/6_conclusion.tex
\section{Conclusion}
\label{sec:conclusion}

In this paper, we introduce \name, a large-scale dataset of dynamic Internet video annotated with camera poses. 
It is curated using carefully designed video filtering and camera pose estimation pipelines, validated through experiments.
We hope \name will enable exciting new possibilities.
For instance, Table~\ref{tab:finetune} shows the dataset can fine-tune DUSt3R~\cite{wang2023dust3r} to produce lower mean error than the synthetic data of MonST3R~\cite{zhang2024monst3r}.
To supplement \name, we collect high-quality synthetic dataset Lightspeed, enabling ground truth pose benchmarking.

\parnobf{Acknowledgments} DF was supported by NSF IIS 2437330. We thank Gabriele Leone and the NVIDIA Lightspeed Content Tech team for sharing the original 3D assets and scene data for creating the Lightspeed benchmark. We thank Yunhao Ge, Zekun Hao, Yin Cui, Xiaohui Zeng, Zhaoshuo Li, Hanzi Mao, Jiahui Huang, Justin Johnson, JJ Park and Andrew Owens for invaluable inspirations, discussions and feedback on this project.

%% file: supp/method_details.tex
\section{DynPose-100K: Dataset Curation Additional Details}
\label{sec:method_details}

We expand upon the method of 
\ifthenelse{\equal{\arxiv}{1}}
{\S~\ref{sec:curation}.}
{\S 3 from the paper.}

\subsection{Candidate Selection Criteria}
\label{sec:method_details_criteria}

Candidate selection criteria fit into three broad categories, which can be broken down into sub-categories. These are visualized in 
\ifthenelse{\equal{\arxiv}{1}}
{Figure~\ref{fig:panda70m_stats}}
{Figure 2 in the main paper}
and are broken down further in Figure~\ref{fig:filtering_supp}. We detail below:

\begin{enumerate}[leftmargin=1.5em,itemsep=0.1em]
    \item[C1.] \textbf{Real-world and quality video.} Videos removed for this reason are titled in Figure~\ref{fig:filtering_supp} not real and quality / ethical. Not real videos are described in 
    \ifthenelse{\equal{\arxiv}{1}}
    {\S~\ref{sec:curation_selection_criteria}}
    {the main paper}
    and include cartoons and animated videos, video games, computer screen recordings, post-processing resulting in large logos or text appearing on screen or other video appearing side-by-side. 
    Quality reasons include poor lighting and blocked or blurred lens.
    We select Panda-70M videos~\cite{chen2024panda} with 720p resolution, but in rare cases a lower-resolution video has been up-sampled to 720p. 
    Ethical reasons include children, NSFW and violence.
    \item[C2.] \textbf{Feasibility for pose prediction.}
    Videos removed for this reason are labeled not estimable. Subcategories include long focal length, zoom in or out, non-existent or out-of-focus static region, shot change, and ambiguous frame of reference.
    \item[C3.] \textbf{Dynamic camera and scene.}
    These correspond to static camera and static scene in Figure~\ref{fig:filtering_supp}.
\end{enumerate}

\subsection{Candidate Video Selection}
\label{sec:method_details_selection}

\input{supp/figures/filtering_supp}

We include videos based on scores coming from the following filters. 
Each filter produces a score between 0 and 1 as to whether the video should be included. 
The average of these is used as the final score, from which we threshold and take all videos with scores higher.
Thresholds for each filter are tuned on a 1K validation set Panda-Val, collected in a similar manner to the 1K Panda-Test.
Validation videos have no overlap with test videos.
Also, long videos containing val videos have no overlap with the long videos containing test videos. 
This is a necessary precaution since Panda-70M videos are clips from long videos in HD-VILA-100M~\cite{xue2022advancing}.

\begin{enumerate}[itemsep=0.1em]
    \item \emph{Cartoons and presenting}. We use the classifiers of Hands23~\cite{cheng2023towards}, which predict whether a video is likely to have interaction; and whether a video contains children, cartoons or screen recordings, or person sitting in front of a camera.
    This uses 4 evenly spaced frames in thumbnail size as input to two classifiers, one for if a video is acceptable, \eg no cartoons, etc.; and one for if interaction is likely to occur, \eg not a static scene. Each predicts a confidence score between 0 and 1. 
    We use a minimum threshold for 0.55 for acceptable and 0.20 for interaction, meaning scores are 1 if greater than a threshold and 0 otherwise.
    Final score is the average of both. 
    \item \emph{Non-perspective distortion}.
    We use DroidCalib~\cite{hagemann2023deep} to predict radial distortion at an interval of 6 fps for efficiency. 
    DroidCalib predicts $\alpha$ from the unified camera model~\cite{mei2007single}. 
    A maximum threshold of 1.00 is used; higher outputs receive score 0 as they are likely distorted.
    \item \emph{Focal length}.
    We use WildCamera~\cite{zhu2023tame} to compute the frame-wise focal length of videos, applying on frames at 6 fps for efficiency.
    Focal lengths are predicted on 720p frames. 
    We apply two variance thresholds. 
    First, we check the difference between 90th and 10th percentile focal lengths.
    If this is greater than 40\% of the mean focal length, it receives sub-score 0; 1 otherwise.
    This is a good check to see if focal length changes over the course of the video, sometimes due to shot change.
    Second, we check variance over a sliding window of 1 second. 
    If focal length changes by more than 20\% over 1 second, it receives sub-score 0 as there is likely zoom or shot change; 1 otherwise. 
    Finally, we apply a mean threshold, giving sub-score 0 if the 80th percentile focal length is greater than 1400; 1 otherwise. 
    This is because extremely long focal lengths are often too focused to clearly see a background, and further have such a small field of view that localization is challenging, even for a human.
    The final scores is the equally weighted average of the three sub-scores.    
    \item \emph{Dynamic object masking}.
    We apply our masking method 
    \ifthenelse{\equal{\arxiv}{1}}
    {(\S~\ref{sec:curation_pose})}
    {(\S 3 from the paper)}
    to compute the dynamic mask size.
    For efficiency, instead of applying masking every 0.5 seconds and propagating forward 0.5 seconds, we apply masking every 1 second and propagate forward 1 second.
    This is useful since propagation is much faster than the other components of masking. 
    Dynamic mask size is computed for each frame, the 90th percentile mask size is then compared to 80\% of the frame size.
    In other words, if the 10\% of frames with biggest masks have a mask greater than 80\% of frame size, the video receives masking score of 0; 1 otherwise.
    \item \emph{Optical flow}.
    We compute the average magnitude of sequential optical flow using RAFT~\cite{teed2020raft}, this is applied at 6 fps.
    Flow bigger than 2.127\% of frame size average distance is given sub-score 1. 
    This threshold is chosen to correspond to the 80th percentile of optical flow in the dataset.
    Low optical flow is removed as it tends to correspond to static scenes and cameras.
    Next, two sub-components handle shot change.
    First, maximum sequential-frame flow must be less than the mean plus 4 standard deviations for 1 on the sub-score; else 0.
    Second, sustained flow must not be too high; maximum average flow over 1 second sliding windows must be less than a threshold of 15\% of frame size to receive 1 on the sub-score; else 0.
    The average of all three components is the final score.
    \item \emph{Point tracking}.
    We apply an abbreviated version of our tracking method over the full video.
    Instead of repeating tracking, we apply tracking only once for efficiency. Tracking is done at 3 fps and tracked for 30 frames, meaning entire videos of 10 seconds or less are tracked fully. From 
    \ifthenelse{\equal{\arxiv}{1}}
    {Figure~\ref{fig:length},}
    {Figure 5 in the main paper,}
    over 99\% of sequences are less than 10 seconds. In the case of longer videos, the first 10 seconds still provides a good estimate of the tracking metrics.
    Tracking has three components, one measuring large disappearance of tracks indicating shot change, and two measuring lack of movement indicating static camera or scene. Each respective formula is more complex, but at a high-level, a video scores 0 on shot change if more than 50\% of tracks are lost across a single frame, while dynamics scores 0 if median tracks move less than about 5\% of frame size.
\end{enumerate}

\parnobf{Component smoothing} A sigmoid is applied to component scores for smoothing. 
While this doesn't impact clear positives or negatives in each case, it gives a more informative ranking of borderline scores.
For example, an example with average optical flow of 2.126\% is marginally less likely to be included than one with average flow of 2.128\%.
Smoothed scoring is more reflective of this rather than the former scoring 0 and the latter scoring 1.

\parnobf{Generalist VLM} The VLM answers eight questions overviewed in 
\ifthenelse{\equal{\arxiv}{1}}
{\S~\ref{sec:curation_filtering}.}
{the main paper.}
Each question indicates a reason a video would not be included, so if any answer is yes, a score of 0 is given. Otherwise, the score is 1. 

\parnobf{Filtering efficiency}
We find we can often reduce computation of filtering methods compared to pose estimation, as not as much precision is required to get an aggregate filtering score.
For example, precisely tracking each frame is important for accurate pose estimation, but to determine whether the camera is static, we need only check if tracks sufficiently moved at a few fps. 
This efficiency is important considering filtering is run on many more videos (about 3.2M) than pose estimation (about 100K).

\parnobf{Collection details}
\label{sec:method_collection_details}
During filtering, we remove videos with average score below a threshold. Final filter scores are between 0 and 1; the final threshold we use is 0.910, resulting in 137K videos, corresponding to about 4.3\% of the full 3.2M videos. 

Filtering is applied in several stages to reduce compute cost, as reported in 
\ifthenelse{\equal{\arxiv}{1}}
{\S~\ref{sec:curation_filtering}.}
{the main paper.}
After completing a stage, only a subset of all filters are available to attempt to remove unsuitable videos. 
Experiments indicate this is not as effective as using all filters (Table~\ref{tab:supp_panda_filter}).
Nevertheless, this subset is still a good proxy for whether to filter the video.
We therefore use an average of these filters, but apply a less strict threshold than the final filter to avoid removing suitable videos.
Subset thresholds are chosen empirically based on the evaluation set to avoid removing good candidates.

Filters are applied in the following order for efficiency: Hands23, flow, focal; then distort; then tracking; then masking; and finally VLM.
Hands23, flow and focal are lightweight operations, distort is nearly as lightweight, then tracking, then masking. 
VLM (GPT-4o~\cite{openai2024gpt4o}) is applied last to minimize OpenAI API costs.
Hands23, flow and focal were run jointly early in the project to experiment with filtering on a large set.

After running SfM, we drop trajectories with less than 80\% of frames registered. 
This is a typical threshold for a succeeded trajectory used in evaluation~\cite{liu2023robust,butler2012naturalistic}. Early experiments found this to be a good proxy for quality.

\subsection{Dynamic Camera Pose Estimation}
\label{sec:method_details_pose}

Dynamic camera pose estimation builds upon ParticleSfM~\cite{zhao2022particlesfm}'s calls to TheiaSfM~\cite{theia-manual} for global bundle adjustment given masks and tracks as input. 
More detail for dynamic masking and point tracking follow.

\parnobf{Dynamic masking} Dynamic masking combines four complimentary approaches to motion detection: semantics, for common dynamic classes; object interaction, for sometimes-dynamic objects that move when interacted with; motion, for sometimes-dynamic objects that move even when not in contact with humans; and tracking, to smoothly and efficiently propagate detected masks.
We apply the former three masks once every six frames and input them into tracking. Tracking then outputs the tracked masks for the current frame and next five frames, before the process is repeated at the sixth new frame. 
Using frames extracted at 12fps for \name, this results in masks produced and propagated forward for 0.5 seconds. 
An example of dynamic masking in practice can be found in more detail in Figure~\ref{fig:supp_masking}.

\input{supp/figures/supp_masking}

\parnoit{Semantic segmentation} We apply OneFormer~\cite{jain2023oneformer} to mask common dynamic classes such as humans and sports equipment. We use the same classes for dynamic masking as RoDynRF~\cite{liu2023robust}: MS-COCO~\cite{lin2014microsoft} classes 1 (person), 2-9 (vehicle), 16-25 (animal), 26-33 (accessory), 34-43 (sports), and 88 (teddy bear).

\parnoit{Object interaction segmentation} We use Hands23~\cite{cheng2023towards} hand-object interaction with default parameters. 
We consider only masks associated with hand-held objects. This corresponds to touch classes 1, 2, 4 and 6.
We found masking objects being touched but not held (3, 5) resulted in masking too many objects that were not moving. Movement was highly correlated with the held class.

\parnoit{Motion segmentation} Sampson error~\cite{hartley2003multiple} is computed both forward and backward on each sequential frame pair based on flow~\cite{teed2020raft}. The maximum of forward and backward is computed for each pixel, after which we normalize and threshold. Like~\cite{liu2023robust}, we mask pixels with errors greater than $(frame height * frame width)/8100$.

\parnoit{Mask propagation} We use SAM2~\cite{ravi2024sam2} video predictor with sam2-hiera-large checkpoint and config to propagate the combination of the three former masks. 
For efficiency, we perform segmentation every six frames and use SAM2 to propagate the mask forward for six frames.
Since propagation is several times faster than masking, the result is a substantial overall speedup.

\parnobf{Point tracking}
To collect \name, we predict tracks using input resolution $(256,256)$, the default for BootsTAP~\cite{doersch2024bootstap}.
We track a grid of $(42,42)=1764$ points.
We apply tracks every 5/12 second and track for 2.5 seconds. 
Using frames extracted at 12fps, this corresponds to a stride of 5 frames and tracking length of 30 frames.
Early experiments indicated higher resolution did not meaningfully improve results while reducing efficiency.
We therefore selected these parameters, considering our goal of large-scale collection.

Repeating tracking later in videos results in sequences where video is shorter than the number of frames to track.
We find BootsTAP~\cite{doersch2024bootstap} runs far faster when repeating the same sequence length, as opposed to shortening sequence length.
We therefore pad sequences with empty frames as needed to keep same sequence length, finding in early experiments this has negligible impact on results while meaningfully speeding up tracking.

%% file: supp/figures/filtering_supp.tex
\begin{figure*}[t]
    \centering
    \includegraphics[width=\linewidth]{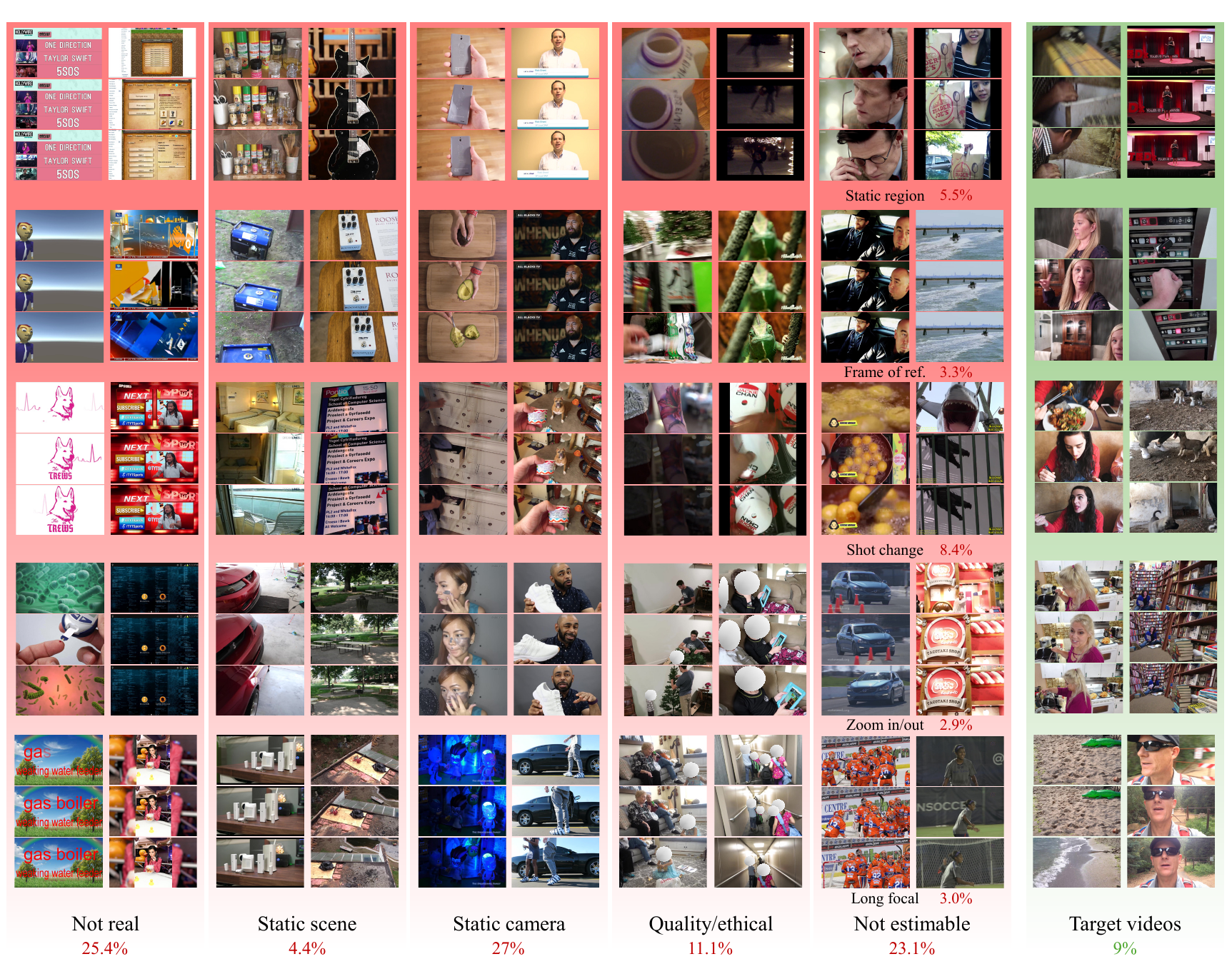}
    \caption{\textbf{Panda-Test breakdown}. We add more examples for each category, including further breakdown statistics for non estimable.
    Stats reflect human labels on the 1K Panda-Test set, detailed in \S~\ref{sec:supp_dataset_details}.
    }
    \label{fig:filtering_supp}
\end{figure*}

%% file: supp/figures/supp_masking.tex
\begin{figure}[t]
    \centering
    \includegraphics[width=\columnwidth]{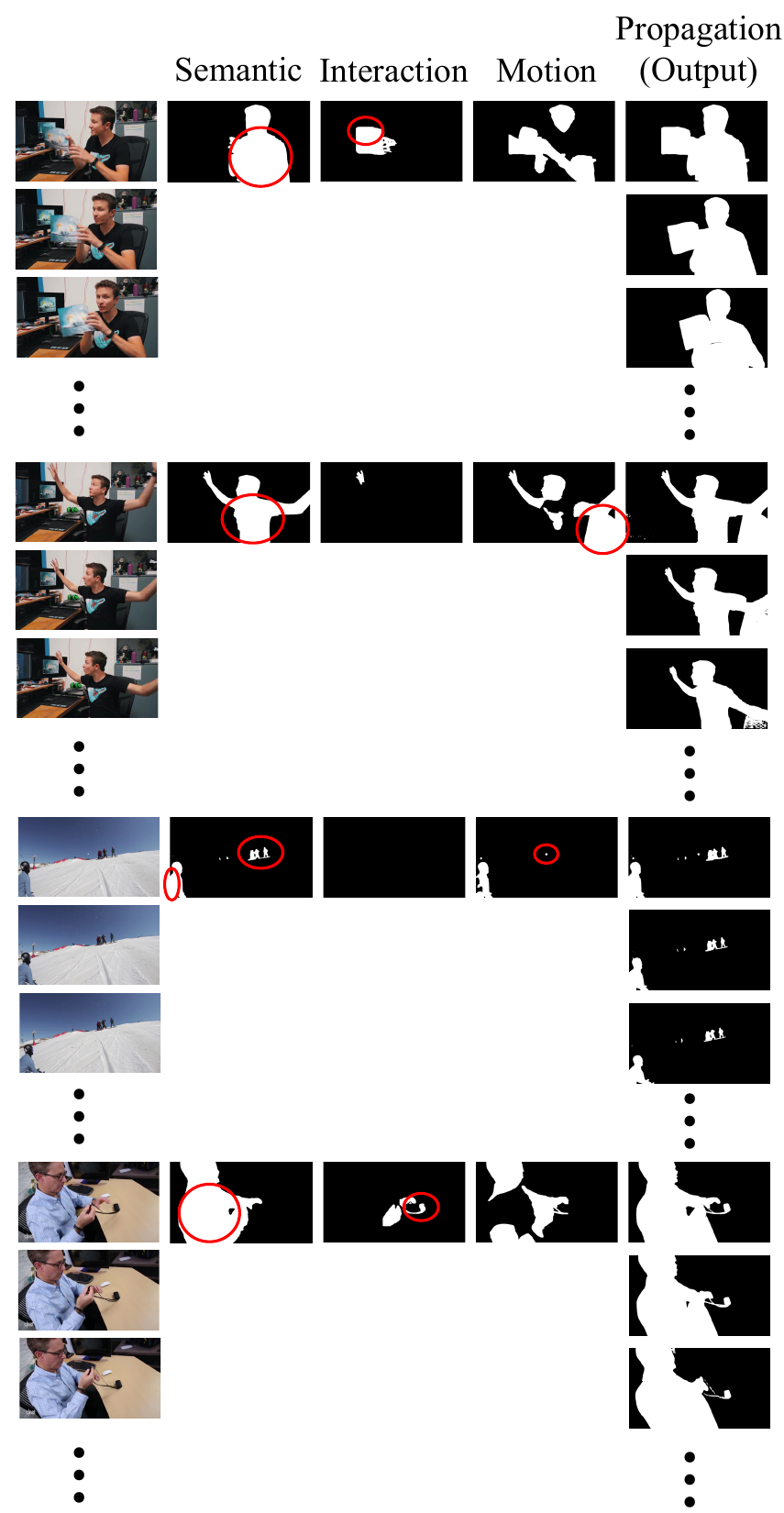}
    \caption{\textbf{Masking composition}. Each component of masking contributes to final masks.
    Unique contributions for semantics, interaction and motion are circled in red.
    Semantic segmentation handles common dynamic objects such as humans (all examples). 
    Object interaction handles things humans are manipulating such as paper (top) or accessories (bottom). 
    Motion handles things moving not by human hands such as swiveling chairs (second from top) or flying snowballs (third from top).
    Sometimes objects are partially segmented by one component but complementary components can still give a more complete mask.
    \Eg in the third example, flying snowballs are segmented by semantics but motion helps complete the dynamic mask.
    All components are combined and tracked smoothly by propagation (right).
    }
    \label{fig:supp_masking}
\end{figure}

%% file: supp/dataset_details.tex
\section{DynPose-100K: Dataset Analysis Additional Details}
\label{sec:experimental_details}

This section adds detail for 
\ifthenelse{\equal{\arxiv}{1}}
{\S~\ref{sec:analysis_stats}.}
{\S 4.2 in the paper.}

\subsection{Filtering Evaluation on Panda-Test}
\label{sec:supp_dataset_details}

\parnobf{Dataset} Panda-Test consists of 1K randomly selected videos manually categorized into suitable or one of several non-suitable categories. 
These are visualized in 
\ifthenelse{\equal{\arxiv}{1}}
{Figure~\ref{fig:panda70m_stats}}
{Figure 2 in the main paper}
and detailed in Figure~\ref{fig:filtering_supp}.
Manual filtering goes as follows: a video is first checked to see if it is not real. 
If it passes this check, it is checked for quality / ethics, then if pose is estimable, then if pose is dynamic, then scene dynamic.
If it passes these checks it is considered a target video.
90 (9\%) are considered suitable. 

\input{supp/tables/full_details_pr}

\parnobf{Results} Table~\ref{tab:supp_panda_filter} shows filtering evaluation in more detail than 
\ifthenelse{\equal{\arxiv}{1}}
{Figure~\ref{fig:panda_filter}.}
{Figure 4 from the main paper.}
Following the Pascal Visual Object Classes challenge~\cite{everingham2010voc}, Table~\ref{tab:supp_panda_filter}, left shows a smoothened precision-recall curve for increased visual clarity. For each data point, we use the maximum precision of all data points whose recall is greater than or equal to the current recall. This smooths the curve such that precision is non-increasing as recall increases. 
Table~\ref{tab:supp_panda_filter}, right reports precision values of the non-smoothened curves.
Precision at Recall of 0.40 corresponds to the operating threshold for \name.

%% file: supp/tables/full_details_pr.tex
\begin{table*}[htp]
\centering
\begin{tabular}{
>{\arraybackslash}m{0.6\textwidth} >{\centering\arraybackslash}m{0.35\textwidth}}

\parbox[c]{\linewidth}{%
    \includegraphics[width=\linewidth]{figures/pdf/pr_baselines_ablation.pdf}
    \captionof{figure}{The \protect\yellowstar represents \name's operating thresholds.
For baselines, we show:
\textcolor{recon_pts}{$\blacksquare$} Reconstructed points (CamCo~\cite{xu2024camco}),
\textcolor{reproj_err}{$\blacksquare$} Reprojection error,
(\textcolor{vlm}{solid $\blacksquare$}) GPT-4o mini~\cite{openai2024gpt4o}: binary,
(\textcolor{vlm}{dashed $\blacksquare$}) GPT-4o mini~\cite{openai2024gpt4o}: score,
\textcolor{hands23}{$\blacksquare$} Hands23~\cite{cheng2023towards}, and
\textcolor{dynpose}{$\blacksquare$} Ours.
For ablations, we begin from \textcolor{hands23}{$\blacksquare$} Hands23 and add components until we recover \textcolor{dynpose}{$\blacksquare$} Ours. Specifically, we depict:
\textcolor{hands23}{$\blacksquare$} Hands23,
\textcolor{flow}{$\blacksquare$} +Flow,
\textcolor{tracking}{$\blacksquare$} +Tracking, 
\textcolor{masking}{$\blacksquare$} +Masking,
\textcolor{focal}{$\blacksquare$} +Focal,
\textcolor{distort}{$\blacksquare$} +Distort,
\textcolor{dynpose}{$\blacksquare$} +VLM (Ours).}
} &

\parbox[c]{\linewidth}{
    \vspace{0in} 
    \setlength{\tabcolsep}{1pt} 
    \resizebox{\linewidth}{!}{ 
    \begin{tabular}{l c c} 
        \toprule
        \multirow{2}{*}{Method} & Avg. & Prec. @ \\
        & Prec. & Rec. $>$0.40$\uparrow$ \\
        \midrule
        Recon. points (CamCo~\cite{xu2024camco}) & 0.17 & 0.19 \\
        Reprojection error       & 0.26 & 0.24 \\
        GPT-4o mini~\cite{openai2024gpt4o}: binary        & 0.18 & 0.24 \\
        GPT-4o mini~\cite{openai2024gpt4o}: score      & 0.13 & 0.13 \\
        Hands23~\cite{cheng2023towards}             & 0.19 & 0.18 \\ 
        Ours             & \textbf{0.58} & \textbf{0.78} \\ 
        \midrule
        Hands23         & 0.19 & 0.18 \\ 
        + Flow          & 0.27 & 0.28 \\ 
        + Tracking      & 0.34 & 0.36 \\
        + Masking       & 0.39 & 0.47 \\
        + Focal         & 0.45 & 0.52 \\ 
        + Distort    & 0.46 & 0.53 \\ 
        + VLM (Ours) & \textbf{0.58} & \textbf{0.78} \\ 
        \bottomrule
    \end{tabular}} 
} \\

\end{tabular}

\ifthenelse{\equal{\arxiv}{1}}
{
    \caption{\textbf{Expanded detail on video filtering on Panda-Test.} 
    Left is Figure~\ref{fig:panda_filter} showing PR curves for baselines and ablations. Right displays average precision for these curves; along with precision at the threshold of 0.40 recall, corresponding to the operating threshold of \name.}
}
{
    \caption{\textbf{Expanded detail on video filtering on Panda-Test.} 
    Left is Figure 4 from the main paper showing PR curves for baselines and ablations. Right displays average precision for these curves; along with precision at the threshold of 0.40 recall, corresponding to the operating threshold of \name.}
}

\label{tab:supp_panda_filter}
\end{table*}

%% file: supp/experimental_details.tex
\section{Experimental Details}
\label{sec:experimental_details}

We expand upon experimental details from
\ifthenelse{\equal{\arxiv}{1}}
{\S~\ref{sec:experiments}.}
{\S 5 in the paper.}

\subsection{Pose Evaluation on Lightspeed}
\label{sec:eval_details_lightspeed}

We run all methods on Lightspeed dataset extracted at full 24 fps due to the short nature of clips, following prior protocol on synthetic data evaluation~\cite{zhao2022particlesfm,liu2023robust,butler2012naturalistic}.

\input{supp/figures/supp_lightspeed_vids}

\parnobf{Dataset}
Additional sequences of Lightspeed are displayed in Figure~\ref{fig:supp_lightspeed_vids}. They have resolution $(2560,1440)$.

\parnobf{Metrics} We follow~\cite{zhao2022particlesfm,liu2023robust,butler2012naturalistic} in defining failing to register a sequence as registering less than 80\% of frames.
We fill failed trajectories with random translations so they can be aligned with ground truth trajectories to compute ATE; \ie the identity sequence cannot be transformed to align with another sequence.
Unlike Panda-Test, we did not find it necessary to repeat SfM:
our method succeeded on all sequences, while repeating SfM on failed sequences for other methods \eg COLMAP made minimal or no difference.

\parnobf{Implementation details}
Our pose estimation method predicts tracks using input resolution $(480,854)$ and a grid of $(56,32)=1792$ points, and tracks for 40 frames. 
This combination is the maximum fitting in 40G memory, and is chosen as a balance of local precision via high-resolution and robustness via long-term tracking.
The 24 fps frame-rate means 40 frames is 1.67 seconds, shorter than the 2.5 seconds on \name and Panda-Test (detailed in \S~\ref{sec:method_details_pose}).
Nevertheless, we find this tracking has competitive final results
\ifthenelse{\equal{\arxiv}{1}}
{(Table~\ref{tab:lightspeed}, Figure~\ref{fig:qualitative_lightspeed},}
{(Table 2, Figure 8 in the main paper,} Figure~\ref{fig:supp_lightspeed_comparison}).
We use tracking applied both forward and backward, though found this had minimal impact on results compared to tracking only forward.

\subsection{Pose Evaluation on Panda-Test}
\label{sec:eval_details_panda}

We use the same settings as \name and apply masking and tracking upon video frames extracted at 12fps.

\parnobf{Metrics}
We select correspondences with small depth (closer to the camera) if possible to make correct pose reprojection error more discernible from incorrect pose estimates.
Sample correspondences used for evaluation are plotted in Figure~\ref{fig:supp_vis_corrs}.
Both human correspondences and corresponding SuperPoint~\cite{detone2018superpoint}+LightGlue~\cite{lindenberger2023lightglue} (SP+LG) correspondences are displayed.
While human correspondences are annotated by two expert annotators and are reliable at a coarse level, SP+LG allows more precise correspondence.
In addition, we require a SP+LG correspondence to exist within 10 pixels (on 720p images) of both human points in correspondence.
This is a fail-safe against a missed annotation from the human.
In practice, we collect 11,866 pairs, about 86.0\% (10,210) of which have agreeing correspondence.
We note SP+LG alone would be difficult to annotate correspondence, since correspondences could potentially belong to a dynamic object or be incorrect.

Non-registered predicted frames are replaced by nearby frames, while non-registered sequences are replaced by the identity matrix.
Both can result in identity relative pose across image pairs.
Identity relative pose results in a singular fundamental matrix, used to compute reprojection error. In this case, instead of reprojection error measuring distance to an epipolar line, distance is computed to the location of the ground truth corresponding point in the opposite image. This can be thought of transforming the correspondence by the identity.
We experimented alternatively adding small random perturbations to enable computation of the fundamental matrix, but this resulted in larger error.

\parnobf{Baselines and ablations}
We find across methods, SfM occasionally fails on challenging Internet video on the same sequence on which it may succeed.
To better analyze performance difference between methods, we therefore repeat failed SfM; defined as registering less than 80\% of frames.

\input{supp/figures/supp_vis_corrs}

\parnobf{Results}
In addition to superior pose estimates, we observe our pose estimation method is faster than ParticleSfM~\cite{zhao2022particlesfm}.
On an A40 GPU, it averages 11 minutes, while ParticleSfM averages 44 minutes, using the same 10 video samples between 48 and 93 frames, with mean length 62.6 (all at 12fps).
The speed difference is a result of ParticleSfM's reliance on propagated dense optical flow for tracking, requiring expensive flow matching to create point tracks. These dense points also result in more correspondences, slowing SfM.
Ours takes about 3.3 minutes for tracking, 4.2 minutes for masking and 3.5 minutes for SfM.
ParticleSfM takes 26 minutes for tracking, 2 for masking and 16 for SfM.
Our quality pose estimates show such dense correspondence is not necessary.
A different setup can provide faster speed across methods. On an A100 40G GPU in a different infra, we saw our inference in as fast as 4 minutes for a video.

On an A100 40G GPU, filtering one video takes about 0.2min for each of Hands23, flow, focal, VLM, and distort; tracking takes 0.8min and masking takes 1.4min. Filtering is performed sequentially to reduce compute and cost.

\parnobf{Fine-tuning experiment}
We compare fine-tuning DUSt3R~\cite{wang2023dust3r} on \name against MonST3R~\cite{zhang2024monst3r}.
To do so, we follow a similar experimental setup to MonST3R, replacing synthetic data used with \name.
For depth supervision, we use pixelwise predictions from Depth-Pro~\cite{Bochkovskii2024}.
We use median depth values to rescale \name translations to meters.
We use only 2K of the 100K videos from \name for this experiment, finding it produced sufficient performance and highlighted supervision efficiency.
We hypothesize using the full dataset could improve performance.
We remove videos not containing all registered frames.
We also remove those with high reprojection error to improve pose accuracy, inspired by Table~\ref{tab:supp_filter_reproj}.
We use a reprojection threshold of 1.37, chosen to provide a balance of quantity and quality at about 1.2K videos for training.
We train for 250K iterations with batch size 8, which takes about 4 days on 2 A40 GPUs.

%% file: supp/figures/supp_lightspeed_vids.tex
\begin{figure*}[t]
    \centering
    \includegraphics[width=\linewidth]{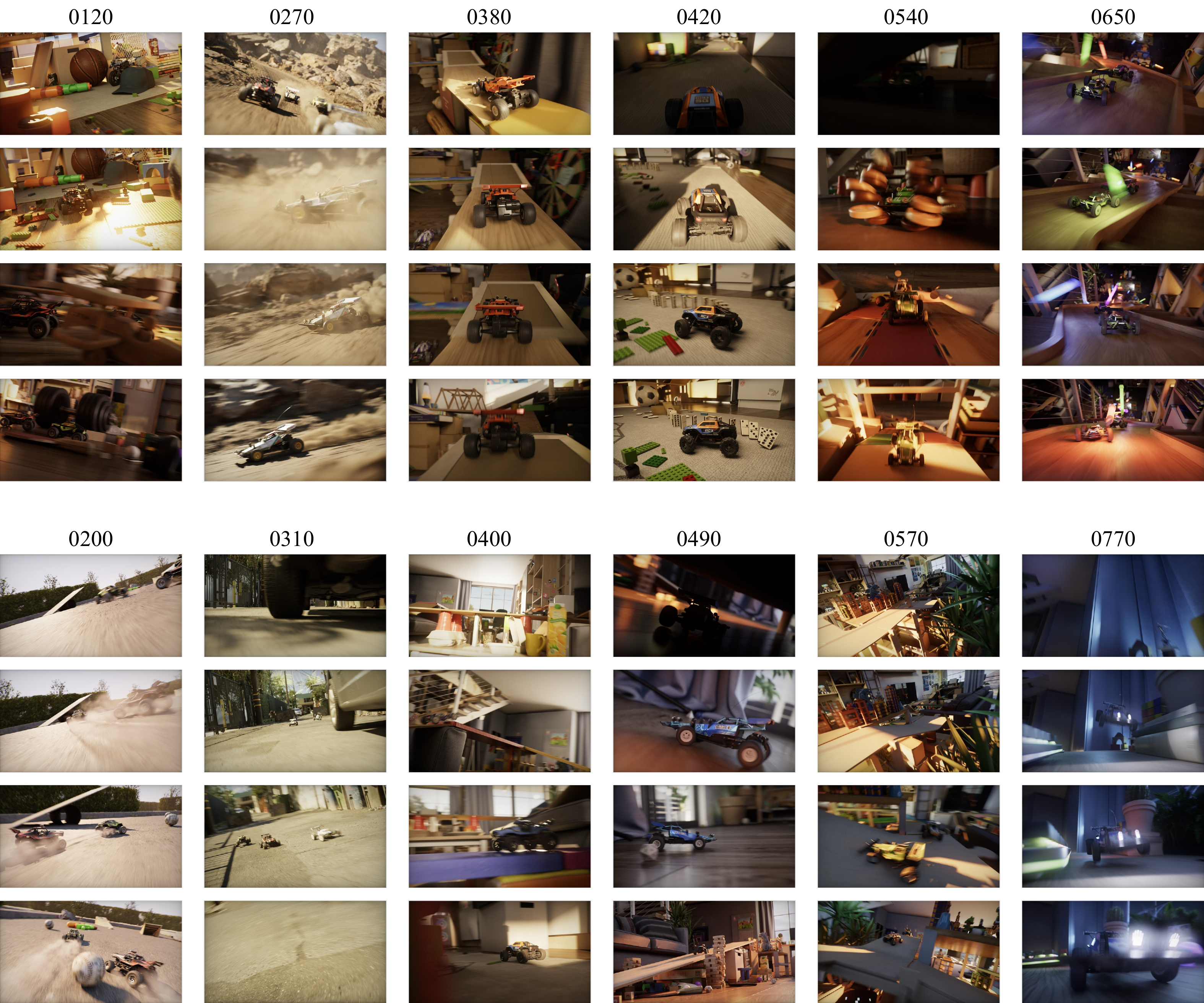}
    \caption{\textbf{Additional Lightspeed videos}. High resolution $(2560,1440)$ sequences span indoor and outdoor; light and dark; close-up dynamic object and far-away dynamic object; forward and backward movement. 
    }
    \label{fig:supp_lightspeed_vids}
\end{figure*}

%% file: supp/figures/supp_vis_corrs.tex
\begin{figure*}[t]
    \centering
    \includegraphics[width=\linewidth]{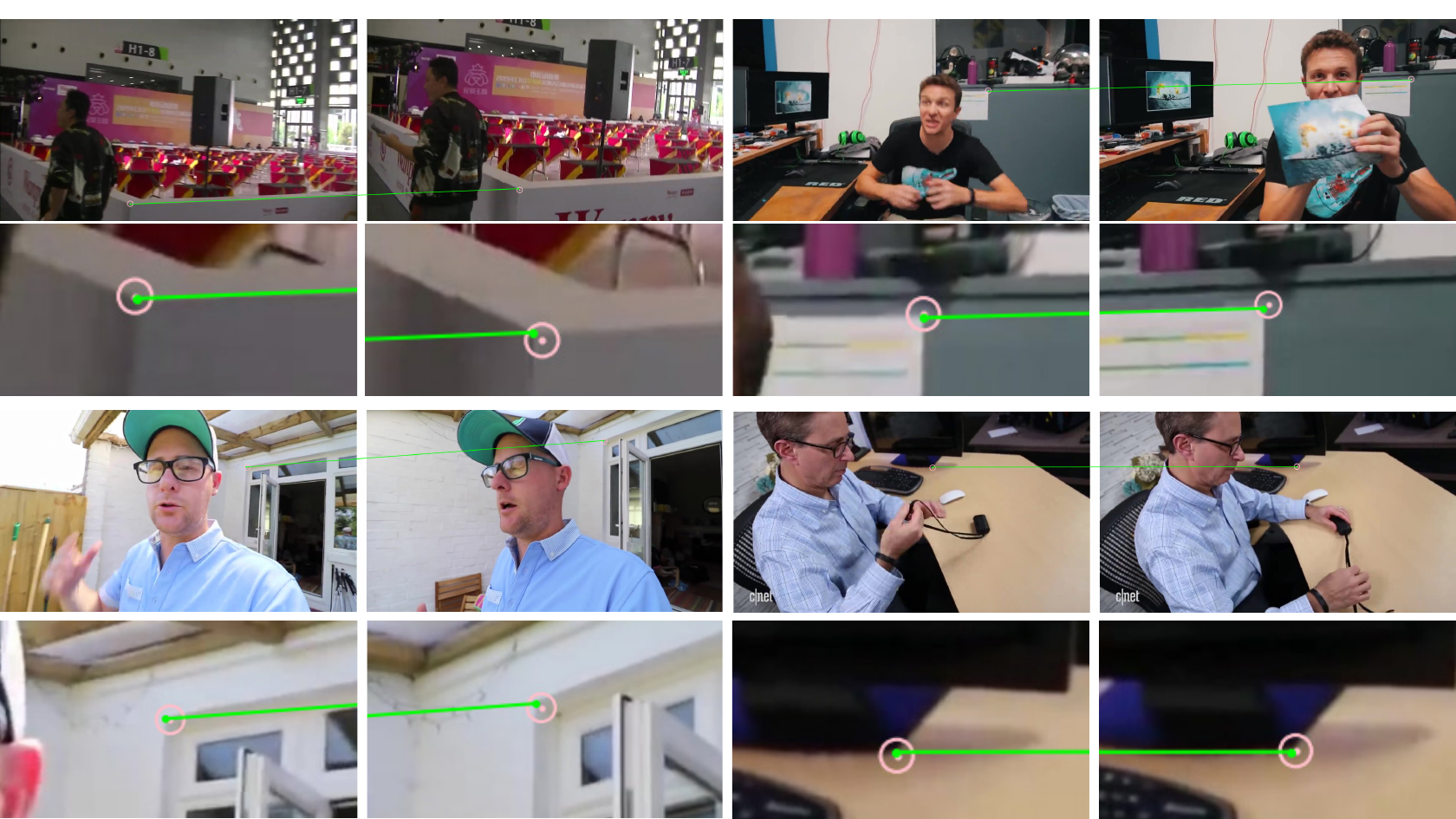}
    \caption{\textbf{Panda-Test correspondence annotation}. Human matches (\textbf{\color{pink}{pink points}}) provide coarse accuracy while SuperPoint+LightGlue~\cite{detone2018superpoint,lindenberger2023lightglue} correspondences (\textbf{\color{green}{green points and line}}) provide precision. 
    We search for SP+LG matches within 10 pixels of human correspondences on 720p frames (\textbf{\color{pink}{pink circle}}). 
    If no match exists, we do not include the match in testing.
    }
    \label{fig:supp_vis_corrs}
\end{figure*}

%% file: supp/additional_dataset_results.tex
\section{DynPose-100K: Dataset Analysis Additional Results}
\label{sec:experimental_details}

We visualize \name filtering process and output.

\subsection{Filtering Evaluation on Panda-Test}
\label{sec:dataset_details}

\input{supp/figures/supp_filtering}

\input{supp/figures/supp_dynpose_100k}

\input{supp/figures/supp_lightspeed_comparison}

Figure~\ref{fig:supp_filtering} breaks down scores for high and low scoring examples on Panda-Test. 
The minimum average score for \name is 0.91, meaning to be included, each filter score must be high. 
The figure shows individual filters can be incomplete, while their combination is typically, but not always, effective:
R1C6 shows the VLM may not catch zoom-in, R2C2 shows Hands will not handle shot change, R2C5 shows masking will not detect static cameras.

\subsection{Dataset Overview}

Figure~\ref{fig:supp_dynpose} displays sample videos and pose annotations on \name. These videos are diverse and face challenges for pose annotation, including varied lighting, movement and apparent dynamic object sizes.
Despite this challenge, pose annotations are of high quality.

\subsection{Reprojection Error Analysis}

Reprojection error is produced during SfM and is saved for each video in DynPose-100K.
It can be used to filter high-quality poses. 
Table~\ref{tab:supp_filter_reproj} reports pose accuracy results of the proposed method before and after filtering by videos with low reprojection error.
Retaining only videos with low reprojection error reduces error significantly.
Future users of DynPose-100K may consider filtering by reprojection error.

\input{supp/tables/filter_reproj}

\subsection{Dataset Rotation Analysis}

Figure~\ref{fig:supp_stats} shows video trajectory rotation statistics. 
Horizontal rotations are particularly diverse, often ending over 30 degrees from the initial rotation with over 75 degrees of total horizontal rotation, measured sequentially.

\input{supp/figures/supp_stats.tex}

%% file: supp/figures/supp_filtering.tex
\begin{figure*}[t]
    \centering
    \includegraphics[width=0.83\linewidth]{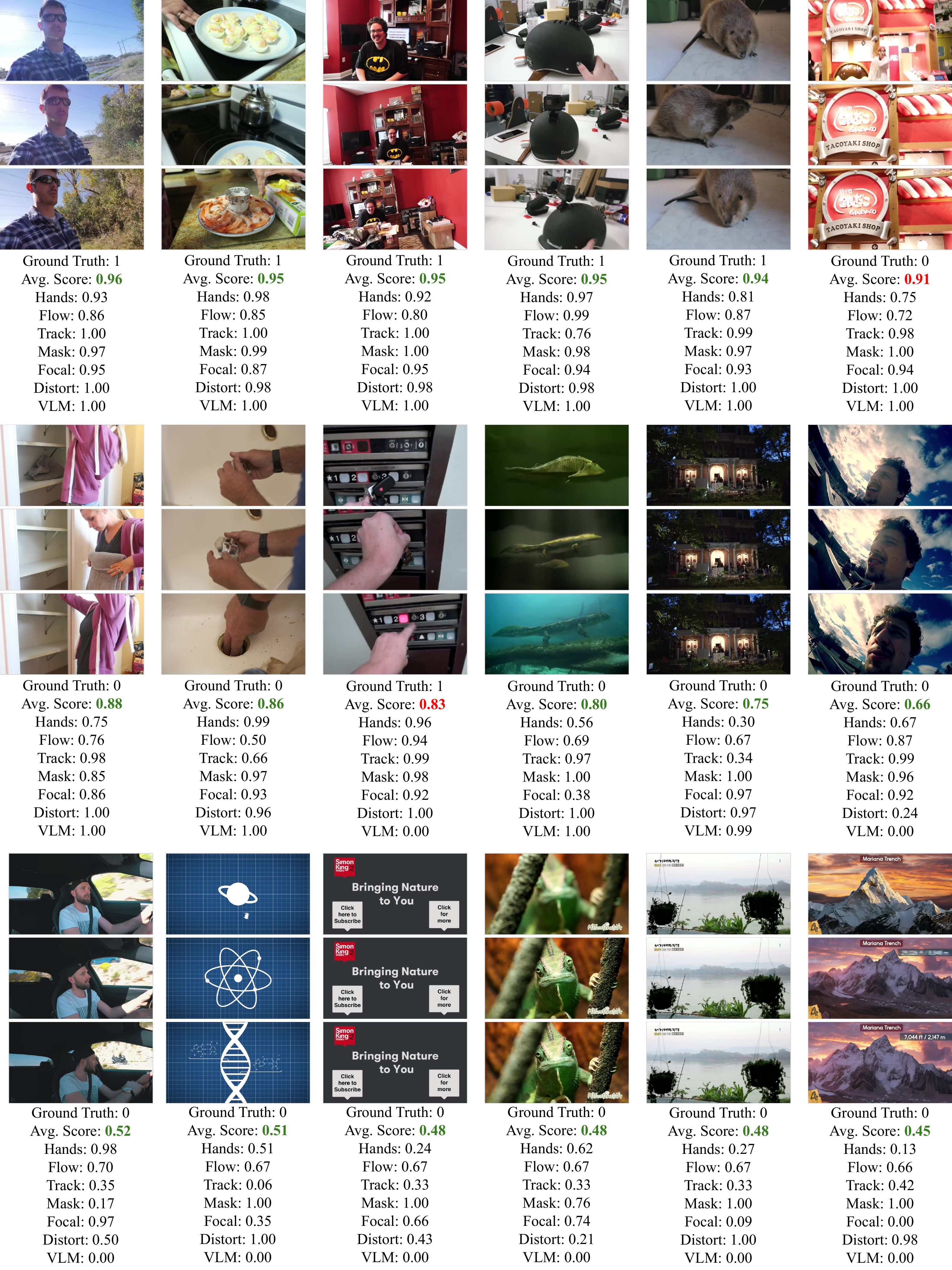}
    \captionof{figure}{\textbf{Panda-Test filtering score samples}. High scoring (top), moderate scoring (middle) and low scoring (bottom) examples from Panda-Test. 
    Ground Truth 1 indicates suitable video, 0 is unsuitable.
    The minimum average score in \name is 0.91, meaning all filters must produce a relatively high score. 
    \textcolor{green}{G} indicates correct classification based on 0.91 threshold; \textcolor{red}{R} are sample failure cases.
    Reasons for exclusion: R1C6: zoom-in, R2C1: static camera, R2C2: shot change, R2C4: not real, R2C5: static camera, R2C6: distortion, R3C1: ambiguous frame of ref, R3C2-C3: not real, R3C4-C5: insufficient clear static region, R2C6: long focal.
    }
    \label{fig:supp_filtering}
\end{figure*}

%% file: supp/figures/supp_dynpose_100k.tex
\begin{figure*}[t]
    \centering
    \includegraphics[width=\linewidth]{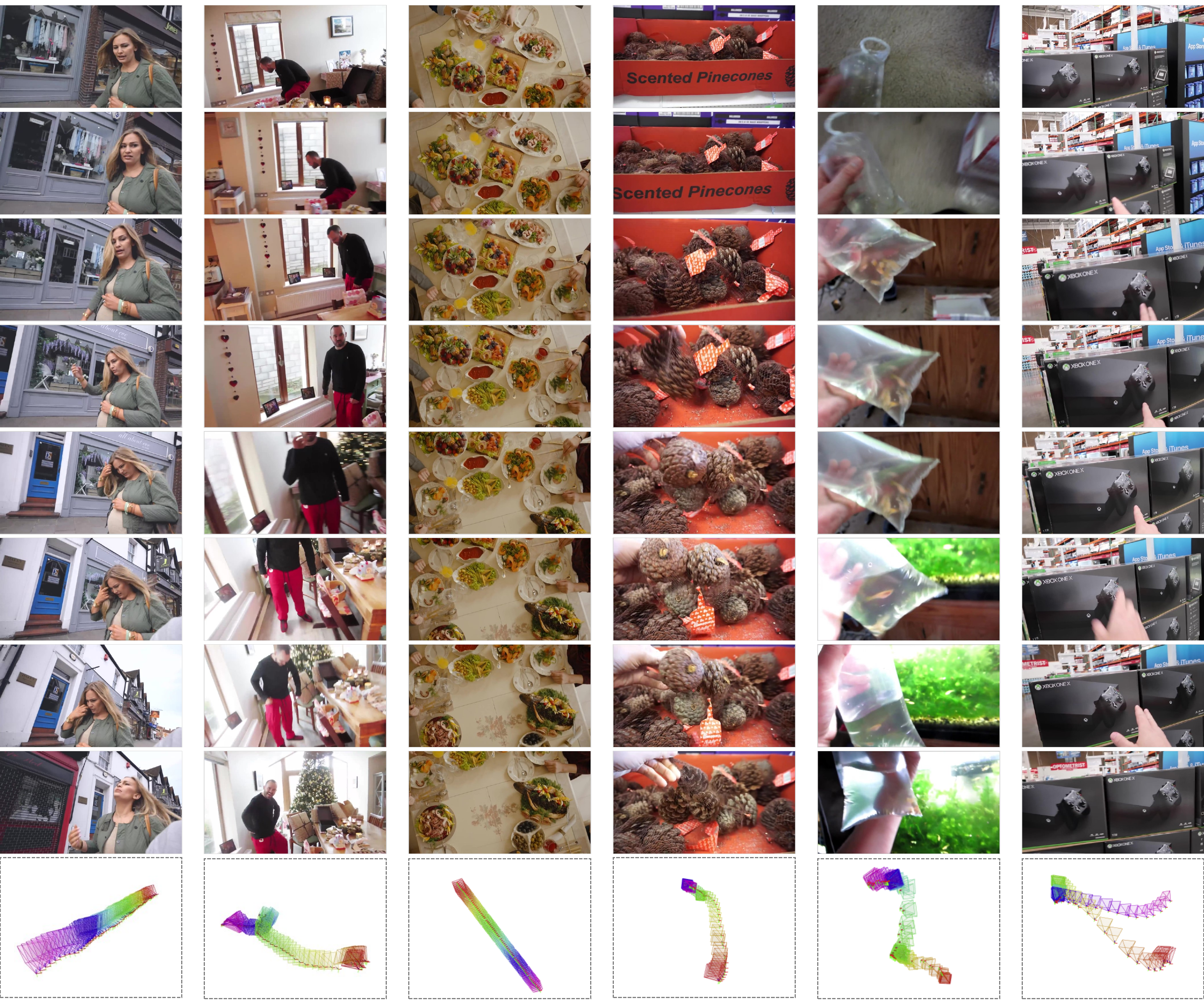}
    \caption{\textbf{Sample videos on \name}. 
    \name collects a diverse set of videos with challenging trajectories and dynamics. It pairs these videos with high-quality pose annotations.
    The dataset is best viewed via the Supplemental video. 
    }
    \label{fig:supp_dynpose}
\end{figure*}

%% file: supp/figures/supp_lightspeed_comparison.tex
\begin{figure*}[t]
    \centering
    \includegraphics[width=\linewidth]{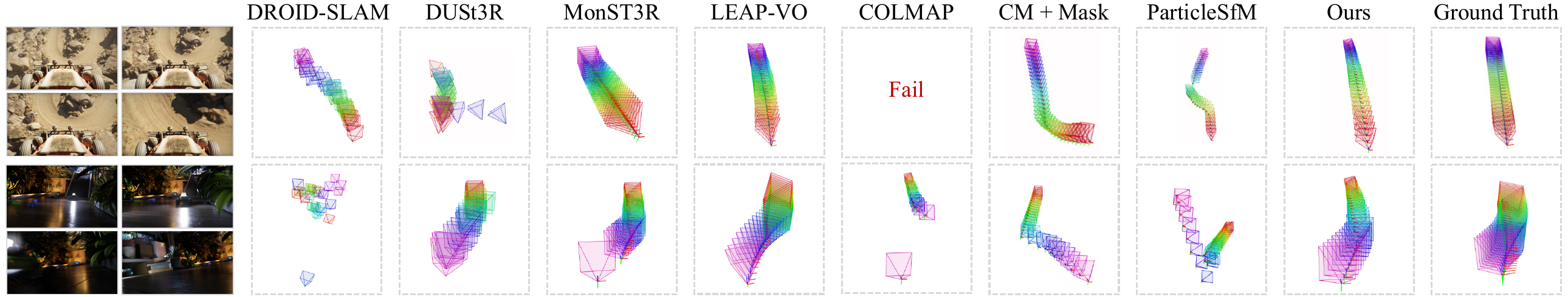}
    \caption{\textbf{Additional comparison on Lightspeed}. 
    Ours has more accurate poses than baselines in challenging settings.
    Top: a dynamic object is static relative to the camera and of similar color to the background. Bottom: dynamic object quickly moves by at night while camera moves and turns. 
    In both cases, baselines are either incorrect or do not have continuous, smooth trajectories.
    Top: MonST3R curves upward towards the end, while COLMAP+Mask has a large turn. 
    }
    \label{fig:supp_lightspeed_comparison}
\end{figure*}

%% file: supp/tables/filter_reproj.tex
\begin{table}
\resizebox{\ifdim\width>\columnwidth \columnwidth \else \width \fi}{!}{
\begin{tabular}{l c c c c} \toprule
Reprojection error & $\%Data\uparrow$ & $\%<5\uparrow$ & $\%<10\uparrow$ & $Mean\downarrow$ \\
\midrule
Full test set & \textbf{100.} & 72.2 & \underline{84.4} & 5.76 \\ 
Reproj. err.$<1.37$ & \underline{71.1} & 78.1 & \underline{84.4} & 5.04 \\
Reproj. err.$<1.18$ & 41.1 & \underline{81.1} & 83.8 & \underline{4.49} \\ 
Reproj. err.$<1.00$ & 24.4 & \textbf{81.8} & \textbf{86.4} & \textbf{3.85} \\ 
\bottomrule
\end{tabular}
}
\caption{\textbf{Identifying high-quality poses.} Reprojection error is effective in identifying low error videos in Panda-Test. It is useful to produce high-quality subsets of DynPose-100K \eg we use reprojection error to help filter data to fine-tune DUSt3R (\S~\ref{sec:eval_details_panda}).}
\label{tab:supp_filter_reproj}
\end{table}

%% file: supp/figures/supp_stats.tex
\begin{table}[t]
\captionsetup{type=figure}
\noindent
\hspace*{-4mm} 
\resizebox{\ifdim\width>\columnwidth \columnwidth \else \width \fi}{!}{
\begin{minipage}{\linewidth}
\begin{tabular}{c c}
        \includegraphics[width=0.51\linewidth]{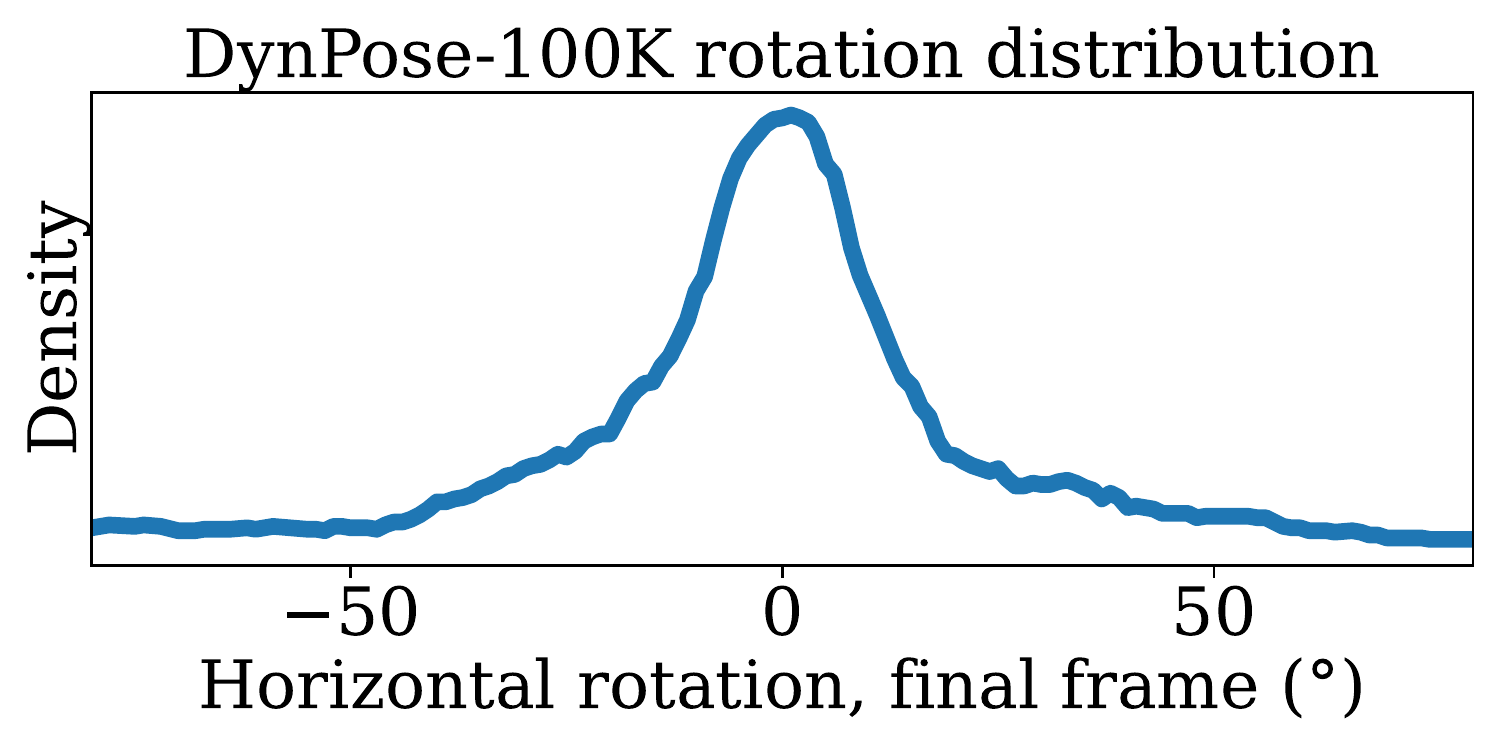}
     &
     \hspace*{-6mm}
        \includegraphics[width=0.51\linewidth]{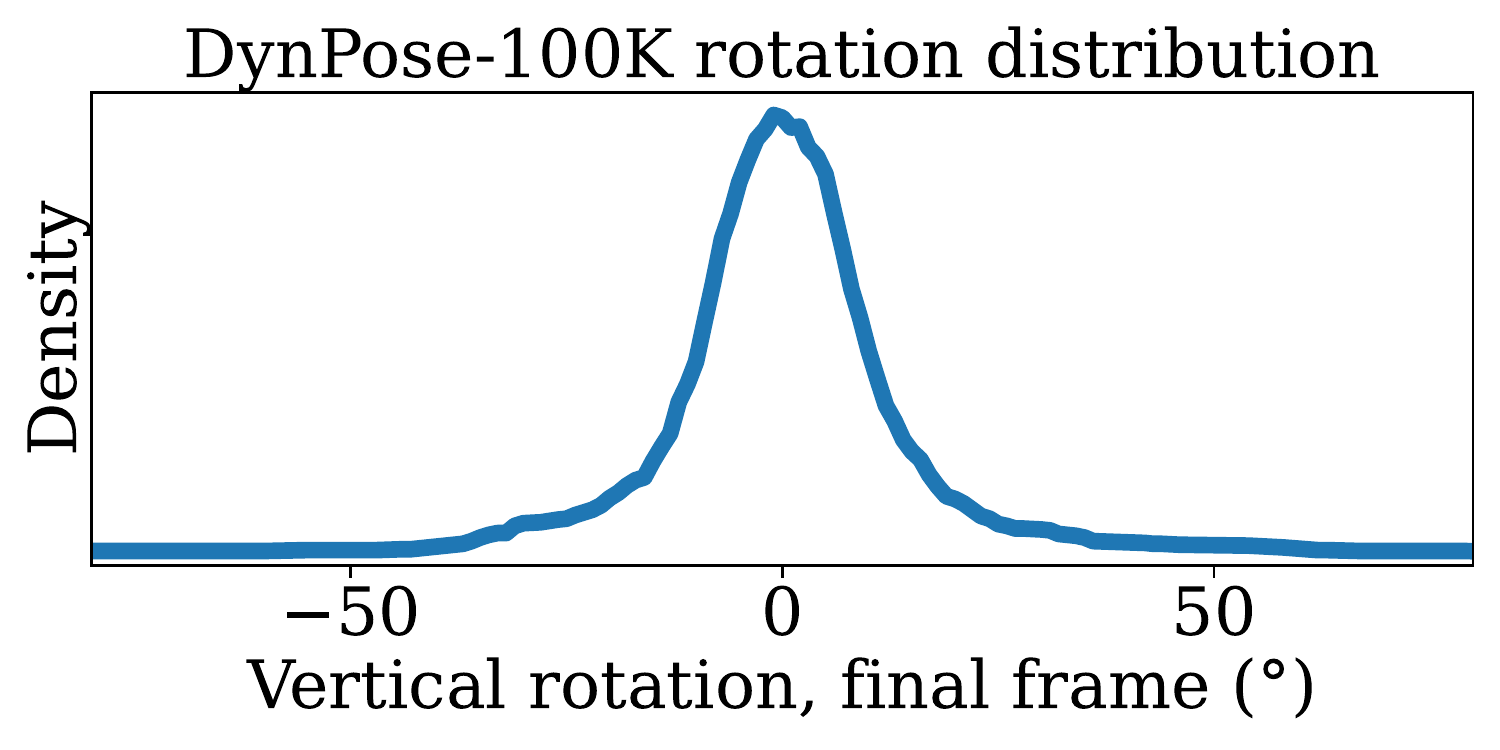}
    \\
    \includegraphics[width=0.51\linewidth]{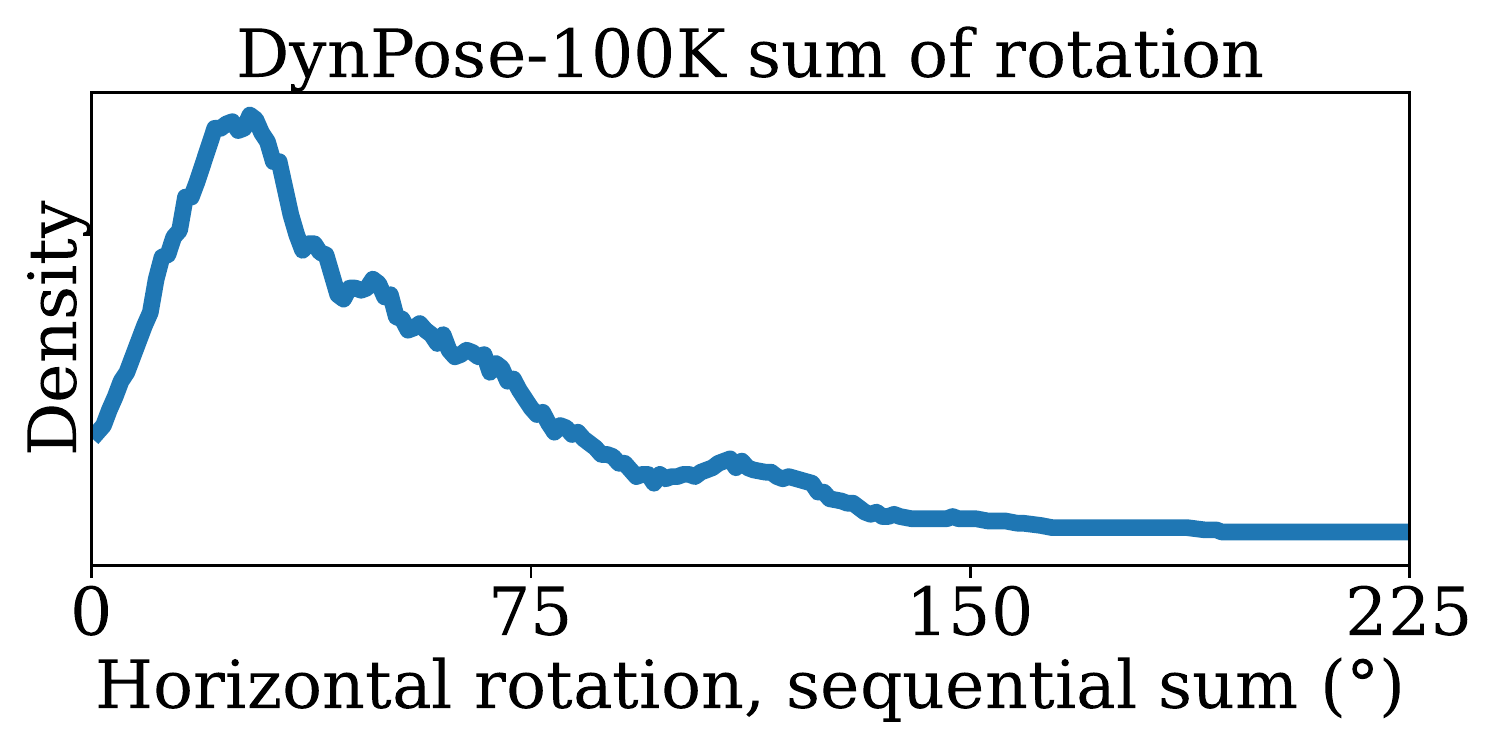}
     &
     \hspace*{-6mm}
        \includegraphics[width=0.51\linewidth]{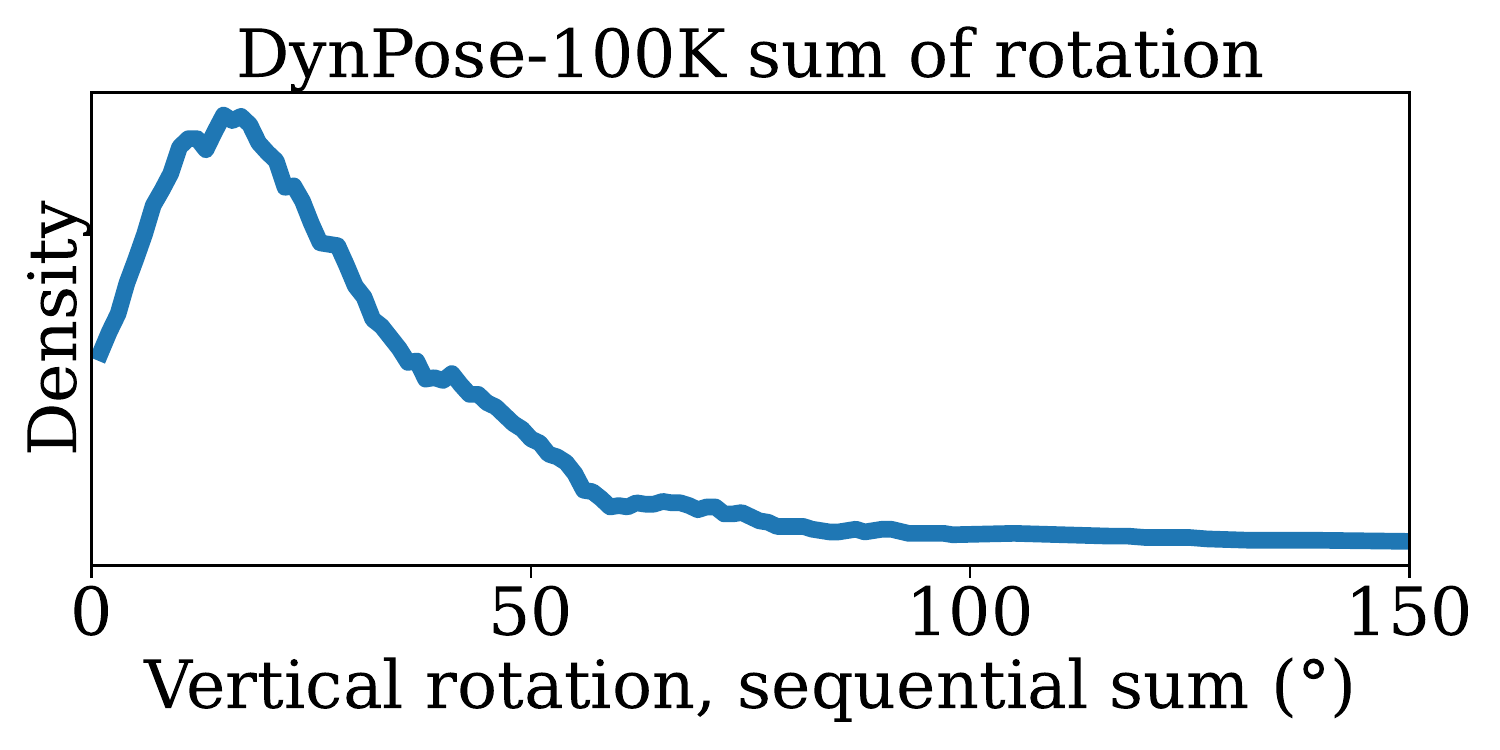}
\end{tabular}
\end{minipage}
}
\caption{\textbf{Dataset statistics.} Top left: distribution of final horizontal rotation. Top right: distribution of final vertical rotation.
Bottom left: distribution of sequential sum of absolute value of horizontal rotation. Bottom right: distribution of sequential sum of absolute value of vertical rotation.
}
\label{fig:supp_stats}
\end{table}

%% file: supp/additional_pose_results.tex
\section{Additional Pose Results}

We show additional results on Lightspeed and Panda-Test.

\subsection{Pose Evaluation on Lightspeed}
\label{sec:exp_lightspeed}

\input{supp/tables/panda70m_tracking_ablations}

Figure~\ref{fig:supp_lightspeed_comparison} displays additional comparisons on Lightspeed. Both examples show all alternatives struggle, while Ours can handle challenges in a large dynamic object that is static relative to the camera (top) and of a dynamic object spinning and kicking up dust (bottom).

\subsection{Pose Evaluation on Panda-Test}
\label{sec:exp_panda}

Figure~\ref{fig:supp_panda_comparison} shows additional comparisons to baselines on Panda-Test. Ours best handles challenges in large dynamic objects and large variations in appearance and lighting. Figure~\ref{fig:supp_panda} shows additional pose results on Panda-Test. Ours produces reasonable trajectories on a wide variety of videos, agreeing with quantitative results.

\parnobf{Tracking ablation} Table~\ref{tab:supp_panda_tracking_ablation} compares our tracking to alternative method CoTracker~\cite{karaev2023cotracker}. We find our tracking produces better pose accuracy.

\parnobf{Masking ablation} Table~\ref{tab:supp_panda_masking_ablations} displays each of the four components to masking. Each component is important to final performance: any component removed reduces results on average.
This is also apparent in Figure~\ref{fig:supp_masking}.

\parnobf{Bundle adjustment ablation} Table~\ref{tab:supp_panda_ba_ablation} compares Ours to alternative bundle adjustment method GLOMAP~\cite{pan2024global}. We find GLOMAP improves precision, but Ours, using Theia-SFM~\cite{theia-manual}, has better mean error.
We note in the GLOMAP paper the method does not outperform Theia-SfM across all scenes. 
Further, the paper measures percentage accuracy within a threshold; our finding on Internet video is GLOMAP failures tend to be more severe than Theia-SfM, which is most clearly reflected in mean error.

\input{supp/tables/panda70m_masking_ablations}

\input{supp/tables/panda70m_ba_ablations}

\input{supp/figures/supp_panda}

\input{supp/figures/supp_panda_comparison}

%% file: supp/tables/panda70m_tracking_ablations.tex
\begin{table}
\resizebox{\ifdim\width>\columnwidth \columnwidth \else \width \fi}{!}{
\begin{tabular}{l c c c c} \toprule
Reprojection error & $\%<5\uparrow$ & $\%<10\uparrow$ & $\%<30\uparrow$ & $Mean\downarrow$ \\
\midrule
Ours & \textbf{72.2} & \textbf{84.4} & \textbf{98.9} & \textbf{5.76} \\ 
+ CoTracker~\cite{karaev2023cotracker} & 66.7 & 78.9 & 97.8 & 7.11 \\ 
\bottomrule
\end{tabular}
}
\caption{\textbf{Tracking ablation on Panda-Test.} CoTracker produces overall worse camera pose estimates than BootsTAP.}
\label{tab:supp_panda_tracking_ablation}
\end{table}

%% file: supp/tables/panda70m_masking_ablations.tex
\begin{table}
\resizebox{\ifdim\width>\columnwidth \columnwidth \else \width \fi}{!}{
\begin{tabular}{l c c c c} \toprule
Reprojection error & $\%<5\uparrow$ & $\%<10\uparrow$ & $\%<30\uparrow$ & $Mean\downarrow$ \\
\midrule
Ours & \textbf{72.2} & 84.4 & \textbf{98.9} & \textbf{5.76} \\ 
- Semantic & 65.6 & \textbf{86.7} & \underline{97.8} & 8.14 \\ 
- Object Interaction & \textbf{72.2} & 85.6 & \underline{97.8} & \underline{5.79} \\ 
- Motion & 67.8 & 81.1 & 94.4 & 7.95 \\ 
- Propagation & 68.9 & \textbf{86.7} & 94.4 & 6.93 \\ 
\bottomrule
\end{tabular}
}
\caption{\textbf{Masking ablations on Panda-Test.} Each component is important to final performance: any component removed reduces results on average.}
\label{tab:supp_panda_masking_ablations}
\end{table}

%% file: supp/tables/panda70m_ba_ablations.tex
\begin{table}

\resizebox{\ifdim\width>\columnwidth \columnwidth \else \width \fi}{!}{
\begin{tabular}{l c c c c} \toprule
Reprojection error & $\%<5\uparrow$ & $\%<10\uparrow$ & $\%<30\uparrow$ & $Mean\downarrow$ \\
\midrule
Ours & 72.2 & 84.4 & \textbf{98.9} & \textbf{5.76} \\ 
+ GLOMAP~\cite{pan2024global} & \textbf{81.1} & \textbf{90.0} & 95.6 & 8.86 \\ 
\bottomrule
\end{tabular}
}
\caption{\textbf{Bundle adjustment ablation on Panda-Test.} GLOMAP offers competitive precision, but Ours has lower mean error.}
\label{tab:supp_panda_ba_ablation}
\end{table}

%% file: supp/figures/supp_panda.tex
\begin{figure*}[t]
    \centering
    \includegraphics[width=\linewidth]{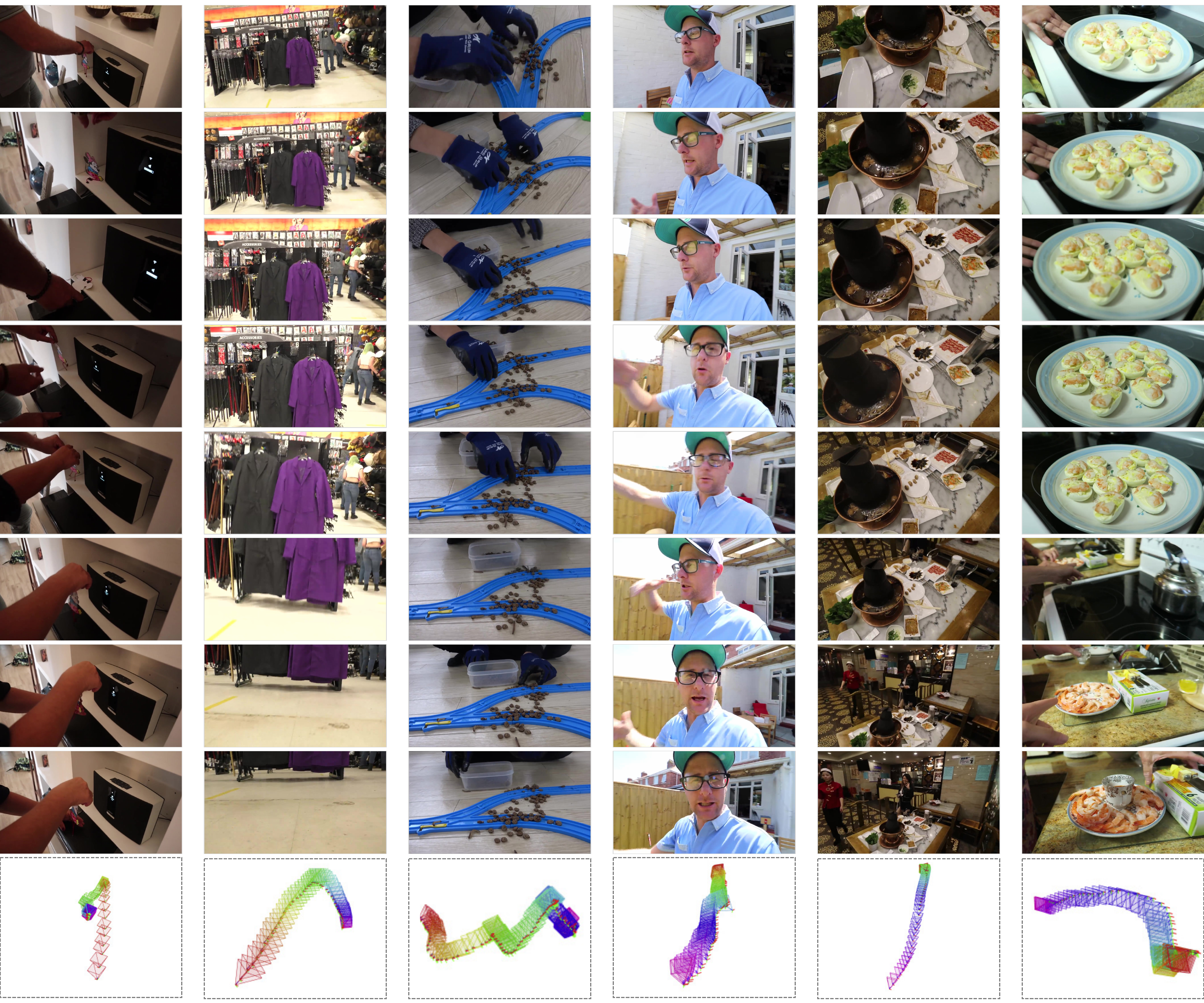}
    \caption{\textbf{Additional poses on Panda-Test}. 
    Ours produces sensible poses on a variety of videos, further validating quantitative results.
    }
    \label{fig:supp_panda}
\end{figure*}

%% file: supp/figures/supp_panda_comparison.tex
\begin{figure*}[t]
    \centering
    \includegraphics[width=\linewidth]{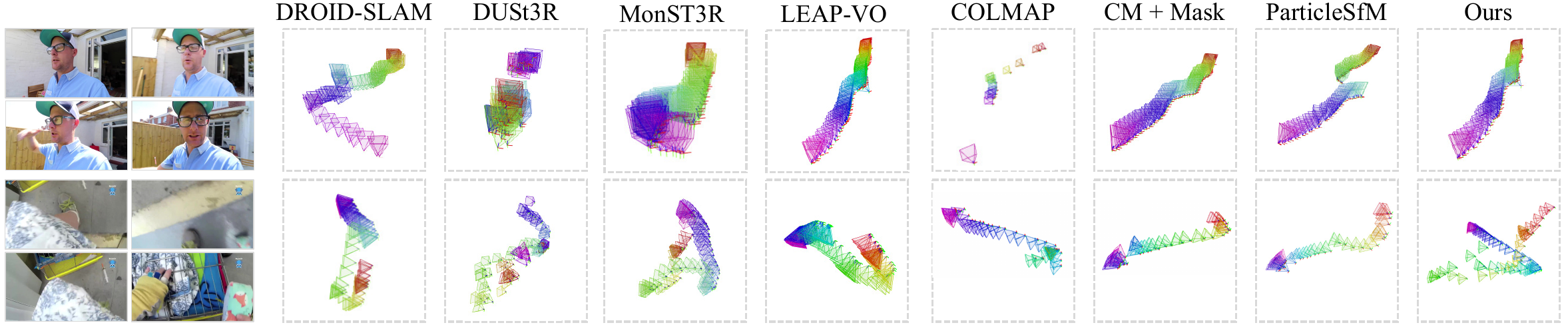}
    \caption{\textbf{Additional comparison on Panda-Test}. 
    Ours best handles challenging lighting and large dynamic objects (top), and handling scale variation resulting from moving very close to objects (bottom). 
    In the top sequence, Our trajectory is accurate, while alternatives do not reflect the consistent, fast and mostly straight backward movement of the camera. 
    COLMAP and COLMAP+Mask register most of the bottom sequence, but miss the movement at the end of the trajectory (down and to the left).
    }
    \label{fig:supp_panda_comparison}
\end{figure*}

%% file: ms.bbl
\begin{thebibliography}{107}
\providecommand{\natexlab}[1]{#1}
\providecommand{\url}[1]{\texttt{#1}}
\expandafter\ifx\csname urlstyle\endcsname\relax
  \providecommand{\doi}[1]{doi: #1}\else
  \providecommand{\doi}{doi: \begingroup \urlstyle{rm}\Url}\fi

\bibitem[Bahl et~al.(2022)Bahl, Gupta, and Pathak]{bahl2022human}
Shikhar Bahl, Abhinav Gupta, and Deepak Pathak.
\newblock Human-to-robot imitation in the wild.
\newblock In \emph{RSS}, 2022.

\bibitem[Bahmani et~al.(2024)Bahmani, Liu, Wang, Skorokhodov, Rong, Liu, Liu,
  Park, Tulyakov, Wetzstein, et~al.]{bahmani2024tc4d}
Sherwin Bahmani, Xian Liu, Yifan Wang, Ivan Skorokhodov, Victor Rong, Ziwei
  Liu, Xihui Liu, Jeong~Joon Park, Sergey Tulyakov, Gordon Wetzstein, et~al.
\newblock Tc4d: Trajectory-conditioned text-to-4d generation.
\newblock In \emph{ECCV}, 2024.

\bibitem[Bain et~al.(2021)Bain, Nagrani, Varol, and Zisserman]{bain2021frozen}
Max Bain, Arsha Nagrani, G{\"u}l Varol, and Andrew Zisserman.
\newblock Frozen in time: A joint video and image encoder for end-to-end
  retrieval.
\newblock In \emph{ICCV}, 2021.

\bibitem[Baruch et~al.(2021)Baruch, Chen, Dehghan, Feigin, Fu, Gebauer, Kurz,
  Dimry, Joffe, Schwartz, et~al.]{baruch1arkitscenes}
Gilad Baruch, Zhuoyuan Chen, Afshin Dehghan, Yuri Feigin, Peter Fu, Thomas
  Gebauer, Daniel Kurz, Tal Dimry, Brandon Joffe, Arik Schwartz, et~al.
\newblock Arkitscenes: A diverse real-world dataset for 3d indoor scene
  understanding using mobile rgb-d data.
\newblock In \emph{NeurIPS}, 2021.

\bibitem[Blattmann et~al.(2023{\natexlab{a}})Blattmann, Dockhorn, Kulal,
  Mendelevitch, Kilian, Lorenz, Levi, English, Voleti, Letts,
  et~al.]{blattmann2023stable}
Andreas Blattmann, Tim Dockhorn, Sumith Kulal, Daniel Mendelevitch, Maciej
  Kilian, Dominik Lorenz, Yam Levi, Zion English, Vikram Voleti, Adam Letts,
  et~al.
\newblock Stable video diffusion: Scaling latent video diffusion models to
  large datasets.
\newblock \emph{arXiv}, 2023{\natexlab{a}}.

\bibitem[Blattmann et~al.(2023{\natexlab{b}})Blattmann, Rombach, Ling,
  Dockhorn, Kim, Fidler, and Kreis]{blattmann2023align}
Andreas Blattmann, Robin Rombach, Huan Ling, Tim Dockhorn, Seung~Wook Kim,
  Sanja Fidler, and Karsten Kreis.
\newblock Align your latents: High-resolution video synthesis with latent
  diffusion models.
\newblock In \emph{CVPR}, 2023{\natexlab{b}}.

\bibitem[Bochkovskii et~al.(2025)Bochkovskii, Delaunoy, Germain, Santos, Zhou,
  Richter, and Koltun]{Bochkovskii2024}
Aleksei Bochkovskii, Ama\"{e}l Delaunoy, Hugo Germain, Marcel Santos, Yichao
  Zhou, Stephan~R. Richter, and Vladlen Koltun.
\newblock Depth pro: Sharp monocular metric depth in less than a second.
\newblock In \emph{ICLR}, 2025.

\bibitem[Brooks et~al.(2024)Brooks, Peebles, Holmes, DePue, Guo, Jing, Schnurr,
  Taylor, Luhman, Luhman, Ng, Wang, and Ramesh]{brooks2024video}
Tim Brooks, Bill Peebles, Connor Holmes, Will DePue, Yufei Guo, Li Jing, David
  Schnurr, Joe Taylor, Troy Luhman, Eric Luhman, Clarence Ng, Ricky Wang, and
  Aditya Ramesh.
\newblock Video generation models as world simulators, 2024.

\bibitem[Butler et~al.(2012)Butler, Wulff, Stanley, and
  Black]{butler2012naturalistic}
Daniel~J Butler, Jonas Wulff, Garrett~B Stanley, and Michael~J Black.
\newblock A naturalistic open source movie for optical flow evaluation.
\newblock In \emph{ECCV}, 2012.

\bibitem[Chan et~al.(2022)Chan, Lin, Chan, Nagano, Pan, De~Mello, Gallo,
  Guibas, Tremblay, Khamis, et~al.]{chan2022efficient}
Eric~R Chan, Connor~Z Lin, Matthew~A Chan, Koki Nagano, Boxiao Pan, Shalini
  De~Mello, Orazio Gallo, Leonidas~J Guibas, Jonathan Tremblay, Sameh Khamis,
  et~al.
\newblock Efficient geometry-aware 3d generative adversarial networks.
\newblock In \emph{CVPR}, 2022.

\bibitem[Chang et~al.(2017)Chang, Dai, Funkhouser, Halber, Niessner, Savva,
  Song, Zeng, and Zhang]{chang2017matterport3d}
Angel Chang, Angela Dai, Thomas Funkhouser, Maciej Halber, Matthias Niessner,
  Manolis Savva, Shuran Song, Andy Zeng, and Yinda Zhang.
\newblock Matterport3d: Learning from rgb-d data in indoor environments.
\newblock In \emph{3DV}, 2017.

\bibitem[Chen et~al.(2024{\natexlab{a}})Chen, Siarohin, Menapace, Deyneka,
  Chao, Jeon, Fang, Lee, Ren, Yang, et~al.]{chen2024panda}
Tsai-Shien Chen, Aliaksandr Siarohin, Willi Menapace, Ekaterina Deyneka,
  Hsiang-wei Chao, Byung~Eun Jeon, Yuwei Fang, Hsin-Ying Lee, Jian Ren,
  Ming-Hsuan Yang, et~al.
\newblock Panda-70m: Captioning 70m videos with multiple cross-modality
  teachers.
\newblock In \emph{CVPR}, 2024{\natexlab{a}}.

\bibitem[Chen et~al.(2024{\natexlab{b}})Chen, Chen, Wang, and
  Pollefeys]{chen2024leap}
Weirong Chen, Le Chen, Rui Wang, and Marc Pollefeys.
\newblock Leap-vo: Long-term effective any point tracking for visual odometry.
\newblock In \emph{CVPR}, 2024{\natexlab{b}}.

\bibitem[Chen et~al.(2024{\natexlab{c}})Chen, Wang, Wang, Wang, and
  Liu]{chen2024v3d}
Zilong Chen, Yikai Wang, Feng Wang, Zhengyi Wang, and Huaping Liu.
\newblock V3d: Video diffusion models are effective 3d generators.
\newblock \emph{arXiv}, 2024{\natexlab{c}}.

\bibitem[Cheng et~al.(2024)Cheng, Oh, Price, Lee, and
  Schwing]{cheng2023putting}
Ho~Kei Cheng, Seoung~Wug Oh, Brian Price, Joon-Young Lee, and Alexander
  Schwing.
\newblock Putting the object back into video object segmentation.
\newblock In \emph{CVPR}, 2024.

\bibitem[Cheng et~al.(2023)Cheng, Shan, Hassen, Higgins, and
  Fouhey]{cheng2023towards}
Tianyi Cheng, Dandan Shan, Ayda Hassen, Richard Higgins, and David Fouhey.
\newblock Towards a richer 2d understanding of hands at scale.
\newblock \emph{NeurIPS}, 2023.

\bibitem[Deitke et~al.(2024)Deitke, Liu, Wallingford, Ngo, Michel, Kusupati,
  Fan, Laforte, Voleti, Gadre, et~al.]{deitke2024objaverse}
Matt Deitke, Ruoshi Liu, Matthew Wallingford, Huong Ngo, Oscar Michel, Aditya
  Kusupati, Alan Fan, Christian Laforte, Vikram Voleti, Samir~Yitzhak Gadre,
  et~al.
\newblock Objaverse-xl: A universe of 10m+ 3d objects.
\newblock In \emph{NeurIPS}, 2024.

\bibitem[DeTone et~al.(2018)DeTone, Malisiewicz, and
  Rabinovich]{detone2018superpoint}
Daniel DeTone, Tomasz Malisiewicz, and Andrew Rabinovich.
\newblock Superpoint: Self-supervised interest point detection and description.
\newblock In \emph{CVPRW}, 2018.

\bibitem[Doersch et~al.(2023)Doersch, Yang, Vecerik, Gokay, Gupta, Aytar,
  Carreira, and Zisserman]{doersch2023tapir}
Carl Doersch, Yi Yang, Mel Vecerik, Dilara Gokay, Ankush Gupta, Yusuf Aytar,
  Joao Carreira, and Andrew Zisserman.
\newblock Tapir: Tracking any point with per-frame initialization and temporal
  refinement.
\newblock In \emph{ICCV}, 2023.

\bibitem[Doersch et~al.(2024)Doersch, Yang, Gokay, Luc, Koppula, Gupta,
  Heyward, Goroshin, Carreira, and Zisserman]{doersch2024bootstap}
Carl Doersch, Yi Yang, Dilara Gokay, Pauline Luc, Skanda Koppula, Ankush Gupta,
  Joseph Heyward, Ross Goroshin, Jo{\~a}o Carreira, and Andrew Zisserman.
\newblock Bootstap: Bootstrapped training for tracking-any-point.
\newblock In \emph{ACCV}, 2024.

\bibitem[Duisterhof et~al.(2024)Duisterhof, Zust, Weinzaepfel, Leroy, Cabon,
  and Revaud]{leroy2024mast3rsfm}
Bardienus Duisterhof, Lojze Zust, Philippe Weinzaepfel, Vincent Leroy, Yohann
  Cabon, and J{\'e}r{\^o}me Revaud.
\newblock Mast3r-sfm: a fully-integrated solution for unconstrained
  structure-from-motion.
\newblock \emph{arXiv}, 2024.

\bibitem[Everingham et~al.(2010)Everingham, Gool, Williams, Winn, and
  Zisserman]{everingham2010voc}
Mark Everingham, Luc Gool, Christopher~K. Williams, John Winn, and Andrew
  Zisserman.
\newblock The pascal visual object classes (voc) challenge.
\newblock \emph{IJCV}, 2010.

\bibitem[Fu et~al.(2024)Fu, Zhao, and Finn]{fu2024mobile}
Zipeng Fu, Tony~Z Zhao, and Chelsea Finn.
\newblock Mobile aloha: Learning bimanual mobile manipulation with low-cost
  whole-body teleoperation.
\newblock In \emph{CoRL}, 2024.

\bibitem[Gao et~al.(2022)Gao, Shen, Wang, Chen, Yin, Li, Litany, Gojcic, and
  Fidler]{gao2022get3d}
Jun Gao, Tianchang Shen, Zian Wang, Wenzheng Chen, Kangxue Yin, Daiqing Li, Or
  Litany, Zan Gojcic, and Sanja Fidler.
\newblock Get3d: A generative model of high quality 3d textured shapes learned
  from images.
\newblock In \emph{NeurIPS}, 2022.

\bibitem[Geyer et~al.(2024)Geyer, Bar-Tal, Bagon, and
  Dekel]{geyer2023tokenflow}
Michal Geyer, Omer Bar-Tal, Shai Bagon, and Tali Dekel.
\newblock Tokenflow: Consistent diffusion features for consistent video
  editing.
\newblock In \emph{ICLR}, 2024.

\bibitem[Goli et~al.(2024)Goli, Sabour, Matthews, Brubaker, Lagun, Jacobson,
  Fleet, Saxena, and Tagliasacchi]{goli2024romo}
Lily Goli, Sara Sabour, Mark Matthews, Marcus Brubaker, Dmitry Lagun, Alec
  Jacobson, David~J Fleet, Saurabh Saxena, and Andrea Tagliasacchi.
\newblock Romo: Robust motion segmentation improves structure from motion.
\newblock \emph{arXiv}, 2024.

\bibitem[Grauman et~al.(2024)Grauman, Westbury, Torresani, Kitani, Malik,
  Afouras, Ashutosh, Baiyya, Bansal, Boote, et~al.]{grauman2023ego}
Kristen Grauman, Andrew Westbury, Lorenzo Torresani, Kris Kitani, Jitendra
  Malik, Triantafyllos Afouras, Kumar Ashutosh, Vijay Baiyya, Siddhant Bansal,
  Bikram Boote, et~al.
\newblock Ego-exo4d: Understanding skilled human activity from first-and
  third-person perspectives.
\newblock In \emph{CVPR}, 2024.

\bibitem[Greff et~al.(2022)Greff, Belletti, Beyer, Doersch, Du, Duckworth,
  Fleet, Gnanapragasam, Golemo, Herrmann, et~al.]{greff2022kubric}
Klaus Greff, Francois Belletti, Lucas Beyer, Carl Doersch, Yilun Du, Daniel
  Duckworth, David~J Fleet, Dan Gnanapragasam, Florian Golemo, Charles
  Herrmann, et~al.
\newblock Kubric: A scalable dataset generator.
\newblock In \emph{CVPR}, 2022.

\bibitem[Hagemann et~al.(2023)Hagemann, Knorr, and Stiller]{hagemann2023deep}
Annika Hagemann, Moritz Knorr, and Christoph Stiller.
\newblock Deep geometry-aware camera self-calibration from video.
\newblock In \emph{ICCV}, 2023.

\bibitem[Harley et~al.(2022)Harley, Fang, and Fragkiadaki]{harley2022particle}
Adam~W Harley, Zhaoyuan Fang, and Katerina Fragkiadaki.
\newblock Particle video revisited: Tracking through occlusions using point
  trajectories.
\newblock In \emph{ECCV}, 2022.

\bibitem[Hartley and Zisserman(2003)]{hartley2003multiple}
Richard Hartley and Andrew Zisserman.
\newblock \emph{Multiple view geometry in computer vision}.
\newblock Cambridge university press, 2003.

\bibitem[He et~al.(2024)He, Xu, Guo, Wetzstein, Dai, Li, and
  Yang]{he2024cameractrl}
Hao He, Yinghao Xu, Yuwei Guo, Gordon Wetzstein, Bo Dai, Hongsheng Li, and
  Ceyuan Yang.
\newblock Cameractrl: Enabling camera control for text-to-video generation.
\newblock \emph{arXiv}, 2024.

\bibitem[He et~al.(2017)He, Gkioxari, Doll{\'a}r, and Girshick]{he2017mask}
Kaiming He, Georgia Gkioxari, Piotr Doll{\'a}r, and Ross Girshick.
\newblock Mask r-cnn.
\newblock In \emph{ICCV}, 2017.

\bibitem[Ho et~al.(2022)Ho, Chan, Saharia, Whang, Gao, Gritsenko, Kingma,
  Poole, Norouzi, Fleet, et~al.]{ho2022imagen}
Jonathan Ho, William Chan, Chitwan Saharia, Jay Whang, Ruiqi Gao, Alexey
  Gritsenko, Diederik~P Kingma, Ben Poole, Mohammad Norouzi, David~J Fleet,
  et~al.
\newblock Imagen video: High definition video generation with diffusion models.
\newblock \emph{arXiv}, 2022.

\bibitem[Hu et~al.(2023)Hu, Russell, Yeo, Murez, Fedoseev, Kendall, Shotton,
  and Corrado]{hu2023gaia}
Anthony Hu, Lloyd Russell, Hudson Yeo, Zak Murez, George Fedoseev, Alex
  Kendall, Jamie Shotton, and Gianluca Corrado.
\newblock Gaia-1: A generative world model for autonomous driving.
\newblock \emph{arXiv}, 2023.

\bibitem[Jain et~al.(2023)Jain, Li, Chiu, Hassani, Orlov, and
  Shi]{jain2023oneformer}
Jitesh Jain, Jiachen Li, Mang~Tik Chiu, Ali Hassani, Nikita Orlov, and Humphrey
  Shi.
\newblock Oneformer: One transformer to rule universal image segmentation.
\newblock In \emph{CVPR}, 2023.

\bibitem[Jin et~al.(2025)Jin, Tucker, Li, Fouhey, Snavely, and
  Holynski]{jin2024stereo4d}
Linyi Jin, Richard Tucker, Zhengqi Li, David Fouhey, Noah Snavely, and
  Aleksander Holynski.
\newblock Stereo4d: Learning how things move in 3d from internet stereo videos.
\newblock In \emph{CVPR}, 2025.

\bibitem[Karaev et~al.(2024)Karaev, Rocco, Graham, Neverova, Vedaldi, and
  Rupprecht]{karaev2023cotracker}
Nikita Karaev, Ignacio Rocco, Benjamin Graham, Natalia Neverova, Andrea
  Vedaldi, and Christian Rupprecht.
\newblock Cotracker: It is better to track together.
\newblock In \emph{ECCV}, 2024.

\bibitem[Kasten et~al.(2024)Kasten, Lu, and Maron]{kasten2024learning}
Yoni Kasten, Wuyue Lu, and Haggai Maron.
\newblock Fast encoder-based 3d from casual videos via point track processing.
\newblock In \emph{NeurIPS}, 2024.

\bibitem[Knapitsch et~al.(2017)Knapitsch, Park, Zhou, and
  Koltun]{knapitsch2017tanks}
Arno Knapitsch, Jaesik Park, Qian-Yi Zhou, and Vladlen Koltun.
\newblock Tanks and temples: Benchmarking large-scale scene reconstruction.
\newblock In \emph{SIGGRAPH}, 2017.

\bibitem[Kopf et~al.(2021)Kopf, Rong, and Huang]{kopf2021robust}
Johannes Kopf, Xuejian Rong, and Jia-Bin Huang.
\newblock Robust consistent video depth estimation.
\newblock In \emph{CVPR}, 2021.

\bibitem[Kuang et~al.(2024)Kuang, Cai, He, Xu, Li, Guibas, and
  Wetzstein]{kuang24}
Zhengfei Kuang, Shengqu Cai, Hao He, Yinghao Xu, Hongsheng Li, Leonidas Guibas,
  and Gordon Wetzstein.
\newblock Collaborative video diffusion: Consistent multi-video generation with
  camera control.
\newblock In \emph{NeurIPS}, 2024.

\bibitem[Li et~al.(2021)Li, Niklaus, Snavely, and Wang]{li2021neural}
Zhengqi Li, Simon Niklaus, Noah Snavely, and Oliver Wang.
\newblock Neural scene flow fields for space-time view synthesis of dynamic
  scenes.
\newblock In \emph{CVPR}, 2021.

\bibitem[Li et~al.(2025)Li, Tucker, Cole, Wang, Jin, Ye, Kanazawa, Holynski,
  and Snavely]{li2024_MegaSaM}
Zhengqi Li, Richard Tucker, Forrester Cole, Qianqian Wang, Linyi Jin, Vickie
  Ye, Angjoo Kanazawa, Aleksander Holynski, and Noah Snavely.
\newblock {MegaSaM}: Accurate, fast and robust structure and motion from casual
  dynamic videos.
\newblock In \emph{CVPR}, 2025.

\bibitem[Liang et~al.(2024)Liang, Ren, Mirzaei, Torralba, Liu, Gilitschenski,
  Fidler, Oztireli, Ling, Gojcic, and Huang]{liang2024btimer}
Hanxue Liang, Jiawei Ren, Ashkan Mirzaei, Antonio Torralba, Ziwei Liu, Igor
  Gilitschenski, Sanja Fidler, Cengiz Oztireli, Huan Ling, Zan Gojcic, and
  Jiahui Huang.
\newblock Feed-forward bullet-time reconstruction of dynamic scenes from
  monocular videos.
\newblock \emph{arXiv}, 2024.

\bibitem[Lin et~al.(2014)Lin, Maire, Belongie, Hays, Perona, Ramanan,
  Doll{\'a}r, and Zitnick]{lin2014microsoft}
Tsung-Yi Lin, Michael Maire, Serge Belongie, James Hays, Pietro Perona, Deva
  Ramanan, Piotr Doll{\'a}r, and C~Lawrence Zitnick.
\newblock Microsoft coco: Common objects in context.
\newblock In \emph{ECCV}, 2014.

\bibitem[Lindenberger et~al.(2023)Lindenberger, Sarlin, and
  Pollefeys]{lindenberger2023lightglue}
Philipp Lindenberger, Paul-Edouard Sarlin, and Marc Pollefeys.
\newblock {LightGlue: Local Feature Matching at Light Speed}.
\newblock In \emph{ICCV}, 2023.

\bibitem[Ling et~al.(2024)Ling, Sheng, Tu, Zhao, Xin, Wan, Yu, Guo, Yu, Lu,
  et~al.]{ling2024dl3dv}
Lu Ling, Yichen Sheng, Zhi Tu, Wentian Zhao, Cheng Xin, Kun Wan, Lantao Yu,
  Qianyu Guo, Zixun Yu, Yawen Lu, et~al.
\newblock Dl3dv-10k: A large-scale scene dataset for deep learning-based 3d
  vision.
\newblock In \emph{CVPR}, 2024.

\bibitem[Liu et~al.(2023{\natexlab{a}})Liu, Wu, Van~Hoorick, Tokmakov,
  Zakharov, and Vondrick]{liu2023zero}
Ruoshi Liu, Rundi Wu, Basile Van~Hoorick, Pavel Tokmakov, Sergey Zakharov, and
  Carl Vondrick.
\newblock Zero-1-to-3: Zero-shot one image to 3d object.
\newblock In \emph{ICCV}, 2023{\natexlab{a}}.

\bibitem[Liu et~al.(2023{\natexlab{b}})Liu, Gao, Meuleman, Tseng, Saraf, Kim,
  Chuang, Kopf, and Huang]{liu2023robust}
Yu-Lun Liu, Chen Gao, Andreas Meuleman, Hung-Yu Tseng, Ayush Saraf, Changil
  Kim, Yung-Yu Chuang, Johannes Kopf, and Jia-Bin Huang.
\newblock Robust dynamic radiance fields.
\newblock In \emph{CVPR}, 2023{\natexlab{b}}.

\bibitem[Mayer et~al.(2016)Mayer, Ilg, Hausser, Fischer, Cremers, Dosovitskiy,
  and Brox]{mayer2016large}
Nikolaus Mayer, Eddy Ilg, Philip Hausser, Philipp Fischer, Daniel Cremers,
  Alexey Dosovitskiy, and Thomas Brox.
\newblock A large dataset to train convolutional networks for disparity,
  optical flow, and scene flow estimation.
\newblock In \emph{CVPR}, 2016.

\bibitem[Mehl et~al.(2023)Mehl, Schmalfuss, Jahedi, Nalivayko, and
  Bruhn]{mehl2023spring}
Lukas Mehl, Jenny Schmalfuss, Azin Jahedi, Yaroslava Nalivayko, and Andr{\'e}s
  Bruhn.
\newblock Spring: A high-resolution high-detail dataset and benchmark for scene
  flow, optical flow and stereo.
\newblock In \emph{CVPR}, 2023.

\bibitem[Mei and Rives(2007)]{mei2007single}
Christopher Mei and Patrick Rives.
\newblock Single view point omnidirectional camera calibration from planar
  grids.
\newblock In \emph{ICRA}, 2007.

\bibitem[Mur-Artal et~al.(2015)Mur-Artal, Montiel, and Tardos]{mur2015orb}
Raul Mur-Artal, Jose Maria~Martinez Montiel, and Juan~D Tardos.
\newblock {ORB-SLAM}: a versatile and accurate monocular {SLAM} system.
\newblock \emph{T-RO}, 2015.

\bibitem[Ni et~al.(2023)Ni, Shi, Li, Huang, and Min]{ni2023conditional}
Haomiao Ni, Changhao Shi, Kai Li, Sharon~X Huang, and Martin~Renqiang Min.
\newblock Conditional image-to-video generation with latent flow diffusion
  models.
\newblock In \emph{CVPR}, 2023.

\bibitem[NVIDIA(2025)]{nvidia2025}
NVIDIA.
\newblock Cosmos world foundation model platform for physical ai.
\newblock \emph{arXiv}, 2025.

\bibitem[{NVIDIA GeForce}({2022})]{nvidia_racer_rtx}
{NVIDIA GeForce}.
\newblock {NVIDIA Racer RTX | The future of graphics powered by GeForce RTX 40
  Series}, {2022}.
\newblock [Online; accessed {Access Date: 2024-11-9}].

\bibitem[O'Neill et~al.(2023)O'Neill, Rehman, Gupta, Maddukuri, Gupta,
  Padalkar, Lee, Pooley, Gupta, Mandlekar, et~al.]{o2023open}
Abby O'Neill, Abdul Rehman, Abhinav Gupta, Abhiram Maddukuri, Abhishek Gupta,
  Abhishek Padalkar, Abraham Lee, Acorn Pooley, Agrim Gupta, Ajay Mandlekar,
  et~al.
\newblock Open x-embodiment: Robotic learning datasets and rt-x models.
\newblock In \emph{CoRL}, 2023.

\bibitem[OpenAI(2023)]{openai2023gpt4}
OpenAI.
\newblock Gpt-4 technical report, 2023.

\bibitem[OpenAI(2024)]{openai2024gpt4o}
OpenAI.
\newblock Gpt-4o mini: advancing cost-efficient intelligence, 2024.

\bibitem[Pan et~al.(2024)Pan, Bar{\'a}th, Pollefeys, and
  Sch{\"o}nberger]{pan2024global}
Linfei Pan, D{\'a}niel Bar{\'a}th, Marc Pollefeys, and Johannes~L
  Sch{\"o}nberger.
\newblock Global structure-from-motion revisited.
\newblock In \emph{ECCV}, 2024.

\bibitem[Qiu et~al.(2022)Qiu, Wang, Wang, Henein, and Scherer]{qiu2022airdos}
Yuheng Qiu, Chen Wang, Wenshan Wang, Mina Henein, and Sebastian Scherer.
\newblock Airdos: Dynamic slam benefits from articulated objects.
\newblock In \emph{ICRA}, 2022.

\bibitem[Ravi et~al.(2024)Ravi, Gabeur, Hu, Hu, Ryali, Ma, Khedr, R{\"a}dle,
  Rolland, Gustafson, Mintun, Pan, Alwala, Carion, Wu, Girshick, Doll{\'a}r,
  and Feichtenhofer]{ravi2024sam2}
Nikhila Ravi, Valentin Gabeur, Yuan-Ting Hu, Ronghang Hu, Chaitanya Ryali,
  Tengyu Ma, Haitham Khedr, Roman R{\"a}dle, Chloe Rolland, Laura Gustafson,
  Eric Mintun, Junting Pan, Kalyan~Vasudev Alwala, Nicolas Carion, Chao-Yuan
  Wu, Ross Girshick, Piotr Doll{\'a}r, and Christoph Feichtenhofer.
\newblock Sam 2: Segment anything in images and videos.
\newblock \emph{arXiv}, 2024.

\bibitem[Reizenstein et~al.(2021)Reizenstein, Shapovalov, Henzler, Sbordone,
  Labatut, and Novotny]{reizenstein2021common}
Jeremy Reizenstein, Roman Shapovalov, Philipp Henzler, Luca Sbordone, Patrick
  Labatut, and David Novotny.
\newblock Common objects in 3d: Large-scale learning and evaluation of
  real-life 3d category reconstruction.
\newblock In \emph{ICCV}, 2021.

\bibitem[Ren et~al.(2025)Ren, Shen, Huang, Ling, Lu, Nimier-David, MÃ¼ller,
  Keller, Fidler, and Gao]{ren2025gen3c}
Xuanchi Ren, Tianchang Shen, Jiahui Huang, Huan Ling, Yifan Lu, Merlin
  Nimier-David, Thomas MÃ¼ller, Alexander Keller, Sanja Fidler, and Jun Gao.
\newblock Gen3c: 3d-informed world-consistent video generation with precise
  camera control.
\newblock In \emph{CVPR}, 2025.

\bibitem[Schonberger and Frahm(2016)]{schonberger2016structure}
Johannes~L Schonberger and Jan-Michael Frahm.
\newblock Structure-from-motion revisited.
\newblock In \emph{CVPR}, 2016.

\bibitem[Schops et~al.(2017)Schops, Schonberger, Galliani, Sattler, Schindler,
  Pollefeys, and Geiger]{schops2017multi}
Thomas Schops, Johannes~L Schonberger, Silvano Galliani, Torsten Sattler,
  Konrad Schindler, Marc Pollefeys, and Andreas Geiger.
\newblock A multi-view stereo benchmark with high-resolution images and
  multi-camera videos.
\newblock In \emph{CVPR}, 2017.

\bibitem[Seidenschwarz et~al.(2024)Seidenschwarz, Zhou, Duisterhof, Ramanan,
  and Leal-Taix{\'e}]{seidenschwarz2024dynomo}
Jenny Seidenschwarz, Qunjie Zhou, Bardienus Duisterhof, Deva Ramanan, and Laura
  Leal-Taix{\'e}.
\newblock Dynomo: Online point tracking by dynamic online monocular gaussian
  reconstruction.
\newblock \emph{arXiv}, 2024.

\bibitem[Shen et~al.(2023)Shen, Cai, Wang, and Scherer]{shen2023dytanvo}
Shihao Shen, Yilin Cai, Wenshan Wang, and Sebastian Scherer.
\newblock Dytanvo: Joint refinement of visual odometry and motion segmentation
  in dynamic environments.
\newblock In \emph{ICRA}, 2023.

\bibitem[Singer et~al.(2022)Singer, Polyak, Hayes, Yin, An, Zhang, Hu, Yang,
  Ashual, Gafni, et~al.]{singer2022make}
Uriel Singer, Adam Polyak, Thomas Hayes, Xi Yin, Jie An, Songyang Zhang, Qiyuan
  Hu, Harry Yang, Oron Ashual, Oran Gafni, et~al.
\newblock Make-a-video: Text-to-video generation without text-video data.
\newblock \emph{arXiv}, 2022.

\bibitem[Singer et~al.(2023)Singer, Sheynin, Polyak, Ashual, Makarov, Kokkinos,
  Goyal, Vedaldi, Parikh, Johnson, et~al.]{singer2023text}
Uriel Singer, Shelly Sheynin, Adam Polyak, Oron Ashual, Iurii Makarov, Filippos
  Kokkinos, Naman Goyal, Andrea Vedaldi, Devi Parikh, Justin Johnson, et~al.
\newblock Text-to-4d dynamic scene generation.
\newblock \emph{arXiv}, 2023.

\bibitem[Sinha et~al.(2023)Sinha, Shapovalov, Reizenstein, Rocco, Neverova,
  Vedaldi, and Novotny]{sinha2023common}
Samarth Sinha, Roman Shapovalov, Jeremy Reizenstein, Ignacio Rocco, Natalia
  Neverova, Andrea Vedaldi, and David Novotny.
\newblock Common pets in 3d: Dynamic new-view synthesis of real-life deformable
  categories.
\newblock In \emph{CVPR}, 2023.

\bibitem[Smith et~al.(2025)Smith, Charatan, Tewari, and
  Sitzmann]{smith2024flowmap}
Cameron Smith, David Charatan, Ayush Tewari, and Vincent Sitzmann.
\newblock Flowmap: High-quality camera poses, intrinsics, and depth via
  gradient descent.
\newblock In \emph{3DV}, 2025.

\bibitem[Snavely(2008)]{snavely08}
Noah Snavely.
\newblock Scene reconstruction and visualization from internet photo
  collections.
\newblock \emph{PhD Thesis}, 2008.

\bibitem[Sun et~al.(2020)Sun, Kretzschmar, Dotiwalla, Chouard, Patnaik, Tsui,
  Guo, Zhou, Chai, Caine, et~al.]{sun2020scalability}
Pei Sun, Henrik Kretzschmar, Xerxes Dotiwalla, Aurelien Chouard, Vijaysai
  Patnaik, Paul Tsui, James Guo, Yin Zhou, Yuning Chai, Benjamin Caine, et~al.
\newblock Scalability in perception for autonomous driving: Waymo open dataset.
\newblock In \emph{CVPR}, 2020.

\bibitem[Sweeney()]{theia-manual}
Chris Sweeney.
\newblock Theia multiview geometry library: Tutorial \& reference.
\newblock \url{http://theia-sfm.org}.

\bibitem[Teed and Deng(2018)]{teed2018deepv2d}
Zachary Teed and Jia Deng.
\newblock Deepv2d: Video to depth with differentiable structure from motion.
\newblock \emph{arXiv}, 2018.

\bibitem[Teed and Deng(2020)]{teed2020raft}
Zachary Teed and Jia Deng.
\newblock Raft: Recurrent all-pairs field transforms for optical flow.
\newblock In \emph{ECCV}, 2020.

\bibitem[Teed and Deng(2021)]{teed2021droid}
Zachary Teed and Jia Deng.
\newblock {DROID-SLAM}: Deep visual {SLAM} for monocular, stereo, and {RGB-D}
  cameras.
\newblock \emph{NeurIPS}, 2021.

\bibitem[Teed et~al.(2023)Teed, Lipson, and Deng]{teed2022deep}
Zachary Teed, Lahav Lipson, and Jia Deng.
\newblock Deep patch visual odometry.
\newblock In \emph{NeurIPS}, 2023.

\bibitem[Tschernezki et~al.(2024)Tschernezki, Darkhalil, Zhu, Fouhey, Laina,
  Larlus, Damen, and Vedaldi]{tschernezki2024epic}
Vadim Tschernezki, Ahmad Darkhalil, Zhifan Zhu, David Fouhey, Iro Laina, Diane
  Larlus, Dima Damen, and Andrea Vedaldi.
\newblock Epic fields: Marrying 3d geometry and video understanding.
\newblock \emph{NeurIPS}, 2024.

\bibitem[Wang et~al.(2024{\natexlab{a}})Wang, Karaev, Rupprecht, and
  Novotny]{wang2023visual}
Jianyuan Wang, Nikita Karaev, Christian Rupprecht, and David Novotny.
\newblock Vggsfm: Visual geometry grounded deep structure from motion.
\newblock In \emph{CVPR}, 2024{\natexlab{a}}.

\bibitem[Wang et~al.(2021{\natexlab{a}})Wang, Wang, Genova, Srinivasan, Zhou,
  Barron, Martin-Brualla, Snavely, and Funkhouser]{wang2021ibrnet}
Qianqian Wang, Zhicheng Wang, Kyle Genova, Pratul~P Srinivasan, Howard Zhou,
  Jonathan~T Barron, Ricardo Martin-Brualla, Noah Snavely, and Thomas
  Funkhouser.
\newblock Ibrnet: Learning multi-view image-based rendering.
\newblock In \emph{CVPR}, 2021{\natexlab{a}}.

\bibitem[Wang et~al.(2023)Wang, Chang, Cai, Li, Hariharan, Holynski, and
  Snavely]{wang2023tracking}
Qianqian Wang, Yen-Yu Chang, Ruojin Cai, Zhengqi Li, Bharath Hariharan,
  Aleksander Holynski, and Noah Snavely.
\newblock Tracking everything everywhere all at once.
\newblock In \emph{ICCV}, 2023.

\bibitem[Wang et~al.(2025)Wang, Zhang, Holynski, Efros, and Kanazawa]{cut3r}
Qianqian Wang, Yifei Zhang, Aleksander Holynski, Alexei~A. Efros, and Angjoo
  Kanazawa.
\newblock Continuous 3d perception model with persistent state.
\newblock In \emph{CVPR}, 2025.

\bibitem[Wang et~al.(2024{\natexlab{b}})Wang, Leroy, Cabon, Chidlovskii, and
  Revaud]{wang2023dust3r}
Shuzhe Wang, Vincent Leroy, Yohann Cabon, Boris Chidlovskii, and Jerome Revaud.
\newblock Dust3r: Geometric 3d vision made easy.
\newblock In \emph{CVPR}, 2024{\natexlab{b}}.

\bibitem[Wang et~al.(2020)Wang, Zhu, Wang, Hu, Qiu, Wang, Hu, Kapoor, and
  Scherer]{wang2020tartanair}
Wenshan Wang, Delong Zhu, Xiangwei Wang, Yaoyu Hu, Yuheng Qiu, Chen Wang, Yafei
  Hu, Ashish Kapoor, and Sebastian Scherer.
\newblock Tartanair: A dataset to push the limits of visual slam.
\newblock In \emph{IROS}, 2020.

\bibitem[Wang et~al.(2021{\natexlab{b}})Wang, Hu, and
  Scherer]{wang2021tartanvo}
Wenshan Wang, Yaoyu Hu, and Sebastian Scherer.
\newblock {TartanVO}: A generalizable learning-based {VO}.
\newblock In \emph{CoRL}, 2021{\natexlab{b}}.

\bibitem[Wang et~al.(2024{\natexlab{c}})Wang, Zhu, Huang, Chen, and
  Lu]{wang2023drivedreamer}
Xiaofeng Wang, Zheng Zhu, Guan Huang, Xinze Chen, and Jiwen Lu.
\newblock Drivedreamer: Towards real-world-driven world models for autonomous
  driving.
\newblock In \emph{ECCV}, 2024{\natexlab{c}}.

\bibitem[Wang et~al.(2024{\natexlab{d}})Wang, Yuan, Wang, Chen, Xia, Luo, and
  Shan]{wang2023motionctrl}
Zhouxia Wang, Ziyang Yuan, Xintao Wang, Tianshui Chen, Menghan Xia, Ping Luo,
  and Ying Shan.
\newblock Motionctrl: A unified and flexible motion controller for video
  generation.
\newblock In \emph{SIGGRAPH}, 2024{\natexlab{d}}.

\bibitem[Wiles et~al.(2020)Wiles, Gkioxari, Szeliski, and
  Johnson]{wiles2020synsin}
Olivia Wiles, Georgia Gkioxari, Richard Szeliski, and Justin Johnson.
\newblock Synsin: End-to-end view synthesis from a single image.
\newblock In \emph{CVPR}, 2020.

\bibitem[Wulff et~al.(2012)Wulff, Butler, Stanley, and
  Black]{Wulff:ECCVws:2012}
J. Wulff, D.~J. Butler, G.~B. Stanley, and M.~J. Black.
\newblock Lessons and insights from creating a synthetic optical flow
  benchmark.
\newblock In \emph{ECCVW}, 2012.

\bibitem[Xu et~al.(2024)Xu, Nie, Liu, Liu, Kautz, Wang, and
  Vahdat]{xu2024camco}
Dejia Xu, Weili Nie, Chao Liu, Sifei Liu, Jan Kautz, Zhangyang Wang, and Arash
  Vahdat.
\newblock Camco: Camera-controllable 3d-consistent image-to-video generation.
\newblock \emph{arXiv}, 2024.

\bibitem[Xue et~al.(2022)Xue, Hang, Zeng, Sun, Liu, Yang, Fu, and
  Guo]{xue2022advancing}
Hongwei Xue, Tiankai Hang, Yanhong Zeng, Yuchong Sun, Bei Liu, Huan Yang,
  Jianlong Fu, and Baining Guo.
\newblock Advancing high-resolution video-language representation with
  large-scale video transcriptions.
\newblock In \emph{CVPR}, 2022.

\bibitem[Yang et~al.(2023)Yang, Zhou, Liu, and Loy]{yang2023rerender}
Shuai Yang, Yifan Zhou, Ziwei Liu, and Chen~Change Loy.
\newblock Rerender a video: Zero-shot text-guided video-to-video translation.
\newblock In \emph{SIGGRAPH Asia}, 2023.

\bibitem[Ye et~al.(2024)Ye, Chen, Zhan, Huang, Huang, Zhu, Bao, Ouyang, He, and
  Zhang]{Ye2024DATAP}
Weicai Ye, Xinyu Chen, Ruohao Zhan, Di Huang, Xiaoshui Huang, Haoyi Zhu, Hujun
  Bao, Wanli Ouyang, Tong He, and Guofeng Zhang.
\newblock Datap-sfm: Dynamic-aware tracking any point for robust dense
  structure from motion in the wild.
\newblock \emph{arxiv}, 2024.

\bibitem[Yeshwanth et~al.(2023)Yeshwanth, Liu, Nie{\ss}ner, and
  Dai]{yeshwanth2023scannet++}
Chandan Yeshwanth, Yueh-Cheng Liu, Matthias Nie{\ss}ner, and Angela Dai.
\newblock Scannet++: A high-fidelity dataset of 3d indoor scenes.
\newblock In \emph{ICCV}, 2023.

\bibitem[Yin and Shi(2018)]{yin2018geonet}
Zhichao Yin and Jianping Shi.
\newblock Geonet: Unsupervised learning of dense depth, optical flow and camera
  pose.
\newblock In \emph{CVPR}, 2018.

\bibitem[Yu et~al.(2021)Yu, Ye, Tancik, and Kanazawa]{yu2021pixelnerf}
Alex Yu, Vickie Ye, Matthew Tancik, and Angjoo Kanazawa.
\newblock pixelnerf: Neural radiance fields from one or few images.
\newblock In \emph{CVPR}, 2021.

\bibitem[Zhang et~al.(2025)Zhang, Herrmann, Hur, Jampani, Darrell, Cole, Sun,
  and Yang]{zhang2024monst3r}
Junyi Zhang, Charles Herrmann, Junhwa Hur, Varun Jampani, Trevor Darrell,
  Forrester Cole, Deqing Sun, and Ming-Hsuan Yang.
\newblock Monst3r: A simple approach for estimating geometry in the presence of
  motion.
\newblock In \emph{ICLR}, 2025.

\bibitem[Zhang et~al.(2022)Zhang, Cole, Li, Rubinstein, Snavely, and
  Freeman]{zhang2022structure}
Zhoutong Zhang, Forrester Cole, Zhengqi Li, Michael Rubinstein, Noah Snavely,
  and William~T Freeman.
\newblock Structure and motion from casual videos.
\newblock In \emph{ECCV}, 2022.

\bibitem[Zhang et~al.(2024)Zhang, Liao, Li, Dai, Qiu, Zhu, Qin, and
  Wang]{zhang2024tora}
Zhenghao Zhang, Junchao Liao, Menghao Li, Zuozhuo Dai, Bingxue Qiu, Siyu Zhu,
  Long Qin, and Weizhi Wang.
\newblock Tora: Trajectory-oriented diffusion transformer for video generation.
\newblock \emph{arXiv}, 2024.

\bibitem[Zhao et~al.(2022)Zhao, Liu, Guo, Wang, and Liu]{zhao2022particlesfm}
Wang Zhao, Shaohui Liu, Hengkai Guo, Wenping Wang, and Yong-Jin Liu.
\newblock Particlesfm: Exploiting dense point trajectories for localizing
  moving cameras in the wild.
\newblock In \emph{ECCV}, 2022.

\bibitem[Zhao et~al.(2025)Zhao, Lin, Lin, Yan, Li, Yang, Wang, Lee, and
  Wang]{zhao2024genxd}
Yuyang Zhao, Chung-Ching Lin, Kevin Lin, Zhiwen Yan, Linjie Li, Zhengyuan Yang,
  Jianfeng Wang, Gim~Hee Lee, and Lijuan Wang.
\newblock Genxd: Generating any 3d and 4d scenes.
\newblock In \emph{ICLR}, 2025.

\bibitem[Zheng et~al.(2023)Zheng, Harley, Shen, Wetzstein, and
  Guibas]{zheng2023pointodyssey}
Yang Zheng, Adam~W Harley, Bokui Shen, Gordon Wetzstein, and Leonidas~J Guibas.
\newblock Pointodyssey: A large-scale synthetic dataset for long-term point
  tracking.
\newblock In \emph{ICCV}, 2023.

\bibitem[Zhou et~al.(2018)Zhou, Tucker, Flynn, Fyffe, and
  Snavely]{zhou2018stereo}
Tinghui Zhou, Richard Tucker, John Flynn, Graham Fyffe, and Noah Snavely.
\newblock Stereo magnification: Learning view synthesis using multiplane
  images.
\newblock In \emph{SIGGRAPH}, 2018.

\bibitem[Zhu et~al.(2023)Zhu, Kumar, Hu, and Liu]{zhu2023tame}
Shengjie Zhu, Abhinav Kumar, Masa Hu, and Xiaoming Liu.
\newblock Tame a wild camera: In-the-wild monocular camera calibration.
\newblock In \emph{NeurIPS}, 2023.

\end{thebibliography}
